%% file: main.tex
\documentclass{article}

\PassOptionsToPackage{numbers,sort&compress}{natbib}
 \usepackage[preprint]{neurips_2026}

\usepackage[utf8]{inputenc} %
\usepackage[T1]{fontenc}    %
\usepackage{hyperref}       %
\usepackage{url}            %
\usepackage{booktabs}       %
\usepackage{amsfonts}       %
\usepackage{amsmath}
\usepackage{derivative}
\usepackage{nicefrac}       %
\usepackage{microtype}      %
\usepackage[svgnames,table]{xcolor}         %
\usepackage{graphicx} %

\usepackage[noabbrev,capitalise]{cleveref}
\usepackage{multirow}
\usepackage[usestackEOL]{stackengine}
\usepackage{etoolbox}        

\usepackage[separate-uncertainty=true]{siunitx}
\usepackage{subcaption}
\usepackage{xspace} %
\usepackage{wrapfig} %
\usepackage{placeins} %
\usepackage{booktabs} %

\newcommand*{\reset}{\ensuremath{\text{reset}}\xspace}

\newcommand*{\spike}{\ensuremath{\text{sp}}\xspace}
\newcommand*{\tspike}{\ensuremath{t_\spike}\xspace}

\newcommand*{\virt}{\ensuremath{\text{virt}}\xspace}

\title{Quadratic integrate-and-fire neurons exhibit less fragmented loss landscapes and outperform leaky integrate-and-fire neurons in spike-based gradient descent}

\author{%
  Carlo~Wenig$^{1,2}$ \\
  \texttt{carlo.wenig@uni-tuebingen.de} \\
  \And
  Raoul-Martin~Memmesheimer$^1$ \\
  \texttt{rm.memmesheimer@uni-bonn.de} \\
  \And
  Christian~Klos$^{1,3}$ \\
  \texttt{cklos@illinois.edu} \\\\
  $^1$ University of Bonn,
  $^2$ University of Tübingen,
  $^3$ University of Illinois Urbana-Champaign
}

\begin{document}

\maketitle

\begin{abstract}
  The ability to train spiking neural networks is essential for modeling biological neural networks as well as for neuromorphic computing. 
  However, for the extensively used leaky integrate-and-fire (LIF) neurons, arbitrarily small parameter changes can induce spike (dis)appearances that disrupt subsequent activity, leading to unstable neural representations and permanently silent neurons during exact spike-based gradient descent.
  Recent work shows that a class of neuron models, which includes the quadratic integrate-and-fire (QIF) neuron, avoids these discontinuities and enables continuous and even smooth spike-based gradient descent. However, it remains unclear whether these advantages translate into practice. Here, we demonstrate that they do so via a controlled comparison between networks of LIF and QIF neurons on the popular Spiking Heidelberg Digits dataset. Specifically, in a first step, we perform a thorough hyperparameter search to optimize both models, revealing a clear performance advantage of QIF neurons. In a second step, we visualize the loss and gradient landscapes. Consistent with their inferior performance, we find that the loss landscapes of LIF neurons, which are discontinuous, appear more fragmented and the related gradients more erratic.
  An analysis of the landscapes of single samples indicates that these features arise from changes in the temporal order of spikes, which often cause disruptive spike (dis)appearances. Overall, our results advocate replacing LIF neurons with neuron models exhibiting continuous spiking dynamics, such as QIF neurons, for gradient descent training.
\end{abstract}

\section{Introduction}
\label{sec:intro}

Spiking neural networks (SNNs) prevail in biological nervous systems and are essential tools in computational neuroscience~\cite{gerstnerNeuronalDynamicsSingle2014,Dayan2001}. Their neurons communicate via brief, stereotypical electrical impulses. This may not be reducible to a communication via spike rates, due to effects that depend on the precise timing of spikes~\cite{Gollisch2008,wolfeSparsePowerfulCortical2010,saalImportanceSpikeTiming2016,soberMillisecondSpikeTiming2018,Xie2024}. Further, SNNs are used in neuromorphic computing, which promises substantial energy savings, in particular for edge applications~\cite{mehonicBraininspiredComputingNeeds2022,Kudithipudi2025}. Across these domains, the ability to train SNNs is critical. 

The dominant approach for training non-spiking neural networks, hereafter referred to as artificial neural networks (ANNs), is gradient descent training~\cite{Goodfellow2016,bishopDeepLearning2024}. It is based on the assumption that loss gradients provide a useful approximation of how the loss changes locally, which is assured by the meaningful differentiability almost everywhere and the continuity of components typically used in all kinds of ANNs. In addition, many key developments in ANN research of the last decades have addressed rugged loss landscapes~\cite{liVisualizingLossLandscape2018, santurkarHowDoesBatch2019} and, related, exploding and vanishing gradients~\cite{pascanu_difficulty_2013}. 

Spiking neuron models, by contrast, appear ill-suited for gradient descent training because they interact via discrete spikes. In the extensively used leaky integrate-and-fire (LIF) neuron model, spikes may (dis)appear at arbitrary points in time due to arbitrarily small parameter changes~\cite{klosSmoothExactGradient2025}, which can induce disruptive changes of subsequent spiking dynamics and discontinuities in the loss landscape. 

For exact, spike-based gradient descent, where spike times are treated as differentiable functions of the learnable parameters~\cite{bohteErrorbackpropagationTemporallyEncoded2002,Eshraghian2023}, the discontinuities are not predictable from gradients computed in their local neighborhood. Hence, the assumption that gradients provide a useful local approximation of the loss can fail. An often encountered consequence is the disappearance of spikes whose existence is important to achieve a small loss~\cite{comsaTemporalCodingSpiking2020,goltzFastEnergyefficientNeuromorphic2021a,mostafaSupervisedLearningBased2017,nowotnyLossShapingEnhances2025,kheradpishehS4NNTemporalBackpropagation2020}. To counteract these consequences, most current approaches rely on ad-hoc measures~\cite{Eshraghian2023}.

Recently, a conceptual advance~\cite{klosSmoothExactGradient2025} has offered a different, more systematic solution. It has discovered that certain neuron models, including the standard quadratic integrate-and-fire (QIF) neuron, exhibit provably continuous or even smooth spiking dynamics and thus loss landscapes. In these models, spikes cannot (dis)appear at arbitrary points in time, but only at the end of a simulation run, where such events do not disrupt subsequent activity. This property also allows the gradient-based addition of new spikes by means of so-called pseudospikes. However, it remains unclear whether this conceptual advance translates to a practical advantage. More broadly, the analysis of loss landscapes for SNNs is still in its infancy.

\subsection{Contributions}
\label{sec:contributions}

\begin{figure}[htb]
    \centering
    \includegraphics[width=\linewidth]{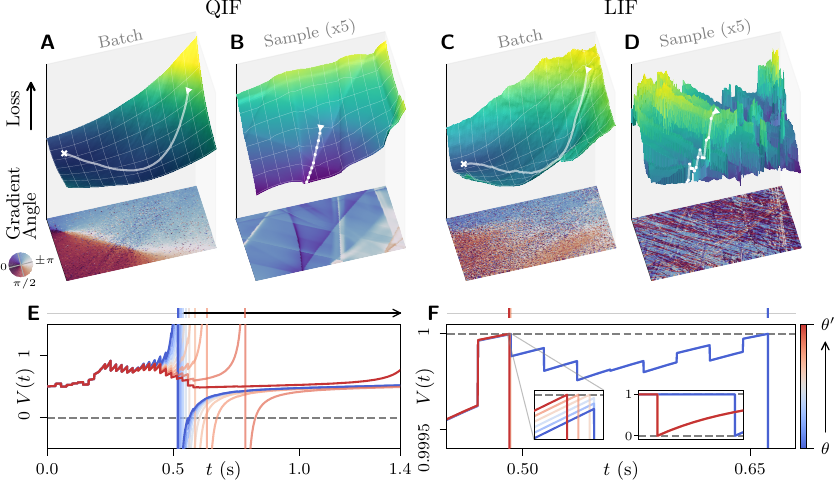}
    \caption{Loss and gradient landscapes of LIF networks are more fragmented than those of QIF networks in the regions of parameter space relevant for learning. 
    (A,C) From a coarse-scale view, batch loss landscapes vary fairly gradually for both QIF and LIF networks, while batch gradient angle landscapes appear more noisy for LIF than QIF networks. (B,D) A fine-scale view of sample landscapes reveals discontinuities in the loss and increased fragmentation of loss and gradient landscapes for LIF networks (wireframe hidden in (D) to decrease clutter). Thick white lines indicate training trajectories (starting at triangles, points show update steps). 
    (E,F) Across fragment boundaries, spike times of some neurons change sharply. 
    Voltage traces (boxes) and spike times (tick marks above) are shown for two example transitions in several steps (intermediate colors) from parameters $\theta$ (blue) to $\theta'$ (red). 
    For the QIF neuron (E), there is initially an early spike. It changes continuously until it leaves the trial at the trial end. For the LIF neuron (F) there is initially a late spike. It changes 
    discontinuously: a spike suddenly appears before the second input spike when the threshold is reached (left inset), while the previously existing spike vanishes due to the new reset (right inset).
    }
    \label{fig:overview}
\end{figure}

We perform a careful comparison between networks of QIF and LIF neurons in a practically relevant setting using exact, spike-based gradient descent. This includes in particular visualization and exploration of the loss and loss gradient landscapes. Specifically, our contributions are as follows:

\begin{itemize}
    \item We perform a thorough hyperparameter search for QIF and LIF networks with infinitesimally short input currents for the Spiking Heidelberg dataset using time-to-first-spike coding. We find that QIF networks outperform LIF networks, particularly when imposing noise-robust distinct output spike times.
    \item We visualize the loss landscape in the region of the training trajectories of the best-performing models. For LIF networks, the sample loss is fragmented into regions bounded by discontinuities. Averaging over samples yields a finer-grained batch loss with attenuated (smaller jump-like changes) but still present discontinuities. For QIF networks, the sample loss is also fragmented, but with continuous, albeit sharp boundaries and to a lesser degree. Averaging over samples yields a visually smooth batch loss. See surface plots in \cref{fig:overview}A-D.
    \item We similarly visualize gradient landscapes. For LIF networks, the sample gradients are even more fragmented and averaging over them does not attenuate the discontinuities, yielding highly erratic batch gradients. For QIF networks, the sample gradients also contain discontinuities, but are much less fragmented, thus yielding less erratic batch gradients. See orientation maps in \cref{fig:overview}A-D.
    \item By zooming into the loss and gradient landscapes, we show that fragment boundaries appear to coincide with changes in the temporal order of spikes. For LIF networks, they often lead to disruptive spike (dis)appearances. For QIF networks, they lead to rapid but continuously changing spike times. See \cref{fig:overview}E,F.
    \item We show that QIF networks exhibit more gradual loss changes between training steps than LIF networks, indicating that the gradient is more predictive of the actual loss change per update. This is consistent with the differences in performance, loss landscapes, and gradient landscapes.
\end{itemize}

\subsection{Related work}
\label{sec:rel_work}

\textbf{Spike-based gradient descent.} Spike-based gradient descent has recently gained popularity due to its use of exact gradients and the ability to exploit sparse communication via spikes during training and simulation~\cite{klosSmoothExactGradient2025,wunderlichEventbasedBackpropagationCan2021a,mullerJaxsnnEventdrivenGradient2024,Konig2026,comsaTemporalCodingSpiking2020,goltzFastEnergyefficientNeuromorphic2021a,mostafaSupervisedLearningBased2017,nowotnyLossShapingEnhances2025,kheradpishehS4NNTemporalBackpropagation2020,engelkenSparsePropEfficientEventBased2023}. Most approaches use LIF or closely related neurons~\cite{wunderlichEventbasedBackpropagationCan2021a,mullerJaxsnnEventdrivenGradient2024,Konig2026,comsaTemporalCodingSpiking2020,goltzFastEnergyefficientNeuromorphic2021a,mostafaSupervisedLearningBased2017,nowotnyLossShapingEnhances2025,kheradpishehS4NNTemporalBackpropagation2020,engelkenSparsePropEfficientEventBased2023}. Spike (dis)appearances are typically dealt with ad-hoc measures such as heuristic changes to input weights or spiking thresholds to revive permanently silent neurons, the introduction of activity regularization, and specialized loss functions~\cite{Eshraghian2023,nowotnyLossShapingEnhances2025}. Inspired by \citet{klosSmoothExactGradient2025}, a few studies have started using QIF neurons with indications of improved performance compared to other neuron models in the context of physics-informed neural networks~\cite{Wan2025} and the development of event-based simulation methods~\cite{Konig2026}.

\textbf{Surrogate gradient descent.} A different approach to gradient-based training of SNNs is surrogate gradient descent~\cite{neftciSurrogateGradientLearning2019a}. It is based on time-discretized SNNs, where the step function, which is necessary for spike detection, is replaced by a continuous-valued surrogate during backpropagation giving rise to non-zero surrogate gradients. This smoothens the jumps in the loss landscapes but induces bias~\cite{Gygax2025} and relies on timestep-based simulations, which are inept to exploit the sparsity of spiking communication.

\textbf{Relation of loss and gradient properties to trainability in ANNs.} 
Many studies have linked loss and gradient properties of ANNs to their trainability. 
Prominent examples have considered residual connections and batch normalization, which enable training deeper networks, and have demonstrated that these techniques smoothen loss landscapes~\cite{liVisualizingLossLandscape2018, santurkarHowDoesBatch2019} and gradients~\cite{balduzzi_shattered_2018, santurkarHowDoesBatch2019}. Other notable examples have shown that flat minima lead to better generalization~\cite{keskarLargeBatchTrainingDeep2016,hochreiter_flat_1997,foretSharpnessAwareMinimizationEfficiently2021} and that minima of ANNs seem to be connected by a path with near constant loss~\cite{garipov_loss_2018,draxler_essentially_2019}.

\textbf{Continuity of spiking dynamics in SNNs.} Only few studies have considered the continuity of spiking dynamics and loss landscapes of time-continuous SNNs. Networks of purely inhibitorily coupled LIF or rapid theta neurons have been shown to exhibit phase spaces that decompose into regions of local stability, termed flux tubes, whose boundaries are related to changes in the temporal order of spikes~\cite{jahnke_stable_2008,monteforteDynamicFluxTubes2012,engelkenSparseChaosCortical2024}. For networks of non-leaky integrate-and-fire neurons that spike at most once, it has been shown that the input domain decomposes into so-called causal pieces, for which output spike times are locally Lipschitz continuous~\cite{doldCausalPiecesAnalysing2025}. 
Finally, \citet{klosSmoothExactGradient2025} have proven the continuity and even smoothness of spiking dynamics and loss landscapes in a class of neuron models that include QIF neurons.

\section{Methods}
\label{sec:methods}

\subsection{Models}
\label{sec:models}

We consider networks of oscillating LIF and QIF neurons. The subthreshold dynamics of an LIF neuron are given by
\begin{align}
\label{eq:LIF voltage}
    \tau \dot V_i(t) = -V_i(t) + I_i(t),
\end{align}
where $i$ is the neuron index, $V_i(t)$ the membrane potential of neuron $i$, $I_i(t)$ the input current of neuron $i$ and $\tau$ the membrane time constant. When $V_i(t)$ surpasses the threshold potential $V_\Theta=1$, neuron $i$ emits a spike and $V_i(t)$ is reset to $V_\reset = 0$. The subthreshold dynamics of a QIF neuron are given by
\begin{align}
\label{eq:QIF voltage}
    \tau \dot V_i(t) = -V_i(t) + V_i^2(t) + I_i(t).
\end{align}
Once the voltage is large enough, the quadratic nonlinearity leads to a voltage self-amplification, analogous to the spike upstroke of biological neurons. The self-amplification is so strong that $V_i(t)$ approaches infinite voltage in finite time. Importantly, this allows to define the time of a spike $\tspike$ as the time when this happens, i.e.\ $V_i(\tspike^-)=\lim_{t\to \tspike^-}V_i(t)=V_\Theta=\infty$. For simplicity, we call $V_\Theta$ the threshold of the QIF neuron. Afterwards, the voltage is reset to negative infinity, i.e.\ $V_i(\tspike^+)=\lim_{t\to \tspike^+}V_i(t)=V_\reset=-\infty$.

For both LIF and QIF neurons, the input current is given by the sum of a constant current $I_0$ and infinitesimally short input currents stemming from other neurons in the network:
\begin{align}
\label{eq:Input current}
    I_i(t) = I_0 + \tau \sum_j w_{ij} \sum_{t_j} \delta(t-t_j),
\end{align}
where $j$ is the index of the presynaptic neurons, which spike at times $t_j$, and $w_{ij}$ are the connection weights. Thus, a spike of neuron $j$ induces an instantaneous jump by $w_{ij}$ in the voltage of neuron $i$. In the case of LIF neurons, this may result in the voltage crossing the threshold potential, after which it will be immediately reset. For QIF neurons, due to the threshold being at infinity, it cannot be reached instantaneously due to an input spike. To ensure that LIF and QIF neurons oscillate intrinsically, $I_0 > V_\Theta = 1$ and $I_0 > \frac{1}{4}$, respectively, are necessary. For such neuron models, there exist simple closed-form solutions for the potential dynamics and for spike. Networks of LIF neurons generically exhibit discontinuous spike times and spike time derivatives, while networks of QIF neurons generically exhibit continuous spike times and discontinuous spike time derivatives~\cite{klosSmoothExactGradient2025}.

To enable the systematic addition of new spikes, we continue to evolve the neurons according to so-called pseudodynamics after a trial has ended~\cite{klosSmoothExactGradient2025}.
At the start of the pseudodynamics, all connected neurons interact with each other. Afterwards, they continue to evolve as in \cref{eq:LIF voltage,eq:QIF voltage,eq:Input current} excluding the interaction between neurons until the neurons have spiked sufficiently often for the considered task.
Pseudodynamics are only used for training and excluded for evaluation. See \cref{sec:model_details} for details.

\subsection{Simulation and gradient computation}
\label{sec:training_sim}

We employ numerically precise event-based simulations~\cite{bretteSimulationNetworksSpiking2007}, iterating over spikes. Briefly, in each iteration, the neuron that spikes next is identified, the corresponding spike time is computed, all neurons are advanced to this time, spike-imposed updates are applied, and the spiking neuron is reset (\cref{sec:simulation_details}). Event-based simulations enable the computation of exact spike-based gradients via automatic differentiation~\cite{klosSmoothExactGradient2025,engelkenSparsePropEfficientEventBased2023,mullerJaxsnnEventdrivenGradient2024,Konig2026} and, unlike previous approaches for the computation of numerically precise exact spike-based gradients~\cite{comsaTemporalCodingSpiking2020,goltzFastEnergyefficientNeuromorphic2021a,mostafaSupervisedLearningBased2017,kheradpishehS4NNTemporalBackpropagation2020}, allow for an arbitrary number of spikes per neuron.

In our simulations, the vast majority of spikes are input spikes. To avoid iterating over all of them, we bin them using periodic virtual input spikes (see code of \cite{klosSmoothExactGradient2025}). Each virtual input spike covers the effect of all input spikes appearing in its neighborhood by summing its output weights, $w_{ik}^\virt = \sum_j w_{ij} n_{jk}$, where $k$ indexes the virtual input spike and $n_{jk}$ is the number of spikes of input neuron $j$ within the $k$-th time bin. This computation is easily parallelized and significantly accelerates the simulation.
For a more detailed description, see \cref{sec:temporal_binning}. While using virtual input spikes discards some temporal information, we find that increasing the temporal resolution beyond 64 bins does not improve classification performance. 
Our code is based on the implementation used by \citet{klosSmoothExactGradient2025} and will be made available upon publication.

\subsection{Task}
\label{sec:task}

We consider the classification of spoken digits using the Spiking Heidelberg Digits (SHD) dataset \cite{cramerHeidelbergSpikingData2022}, which is a standard benchmark task for SNNs. The SHD dataset consists of audio samples of the digits zero to nine in English and German, which are converted into spike trains of 700 input neurons using a model of the inner ear. Each of the 10{,}420 samples contains between 2{,}410 and 14{,}917 spikes.

We train two-layer feedforward networks with 128 hidden neurons and 20 output neurons, one for each class, using batches of 1{,}000 samples, which are reshuffled in every epoch. The input data is preprocessed and augmented using established methods (\cref{sec:data_preprocessing}).

We use time-to-first-spike coding, i.e.\ there is one output neuron per class and the predicted class is given by the first spiking neuron. Such coding is widely used in neuromorphic computing~\cite{wunderlichEventbasedBackpropagationCan2021a,goltzFastEnergyefficientNeuromorphic2021a,cramerHeidelbergSpikingData2022,nowotnyLossShapingEnhances2025} and supported by experimental evidence from neuroscience~\cite{Gollisch2008,Johansson2004,thorpeSpikebasedStrategiesRapid2001}. For training, we use the standard cross-entropy loss with logits $-t_i^\text{first}/T$, where $t_i^\text{first}$ is the time of the first spike of output neuron $i$, which may be either an ordinary spike or a pseudospike, and $T$ is the trial length. 
The logits are scaled by the logit scale $\beta$, also commonly referred to as the inverse temperature. 
To ensure the model does not rely on pseudospikes for classification after training, we add a regularization term that pulls the first spike of the neuron coding for the correct output class forward in time in case it is a pseudospike (\cref{sec:model_details}). 

\section{Results}
\label{sec:results}

\subsection{Model performance}
\label{sec:model_performance}

\begin{figure}[htb]
    \centering
    \includegraphics[width=\linewidth]{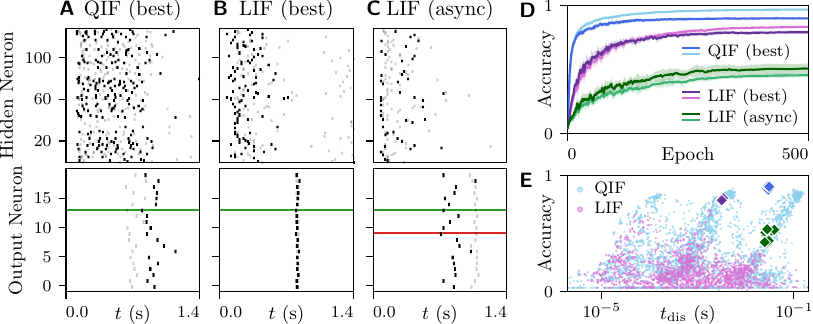}
    \caption{Comparison of the three selected models, the best-performing QIF model, the best-performing LIF model and the asynchronous LIF model. 
    (A-C) Example spike trains before (gray) and after (black) training. True label indicated by green line, model prediction by red line, if different from true label.
    (D) Accuracy during training for the training (brighter colors) and test (darker colors) datasets. Solid lines indicate mean and shaded area standard deviation over ten seeds. 
    (E) Test accuracy vs.\ average distinctness in final epoch of the rough hyperparameter search configurations.
    The three selected models are indicated by diamonds colored as in D. Only QIF networks achieve both high distinctness and high accuracy.}
    \label{fig:model_comparison}
\end{figure}

For a fair comparison of LIF and QIF networks, we perform a thorough hyperparameter search. 
Specifically, we first perform a rough grid search and then an additional finer grid search focusing on the region with highest test accuracies (see \cref{sec:hparam_search} for details including the performance for each parameter set and \cref{tab:hparams} for the resulting final hyperparameters). 
Afterwards, we train both models again for 500 epochs with 10 different seeds. 
For the best-performing QIF model, we find a final test accuracy of \qty{90.1(9)}{\%}, compared to only \qty{79.2(13)}{\%} for LIF neurons (\cref{fig:model_comparison}D).
In addition, the LIF model exhibits much more synchronous output spikes after training compared to the QIF model (\cref{fig:model_comparison} A,B). This is because the hyperparameter search yields a comparatively small initial weight scale and learning rate and a large logit scale for the LIF model, resulting in a small initial temporal spread of the output spikes and small changes of output spike times during training.

Given that almost synchronous output spikes are highly susceptible to noise, which is present in many neuromorphic hardware systems~\cite{Roy2019}, we also seek an LIF model with asynchronous output spikes. 
To this end, we measure the difference between the first spike times of the output neurons that spike first and second, averaged over all samples in the training set. We call this distinctness $t_\text{dis}$  (\cref{fig:model_comparison}E).
The best-performing LIF model has a low average distinctness of \qty{3}{ms} in the final epoch (averaged over samples and seeds), while the best-performing QIF model has one of \qty{28}{ms}. 
There is only a single LIF hyperparameter configuration in the fine search with an average distinctness above \qty{20}{ms}, which we select for the asynchronous model (\cref{fig:model_comparison}C, hyperparameters in \cref{tab:hparams}) and also train again for 500 epochs and 10 seeds. While having an average distinctness of \qty{29}{ms} in the final epoch, comparable to the best QIF model, this LIF network only achieves about \qty{50.8(40)}{\%} test accuracy (\cref{fig:model_comparison}D,E).

In additional exploratory experiments with larger networks, we find an improved final test accuracy of \qty{91.3(8)}{\%} for a QIF network and \qty{85.2(13)}{\%} for an LIF network (see \cref{sec:large_network_results}). While these results are still markedly below the state-of-the-art for the SHD dataset of about 96\% \cite{sunParameterfreeAttentionalSpiking2025,schoneScalableEventbyeventProcessing2024a,baronigAdvancingSpatioTemporalProcessing2025} test accuracy, we note that we use comparably simple network architectures and neurons and focus on the comparison between LIF and QIF neurons rather than maximizing accuracy.

\subsection{Loss and gradient landscapes}
\label{sec:landscapes}

To investigate the origin of the performance difference between QIF and LIF models, we visualize the loss and gradient landscapes of all three selected models.
Due to the high dimensionality of the parameter space, it is impossible to visualize the landscapes in their entirety. We thus select a slice of a two-dimensional plane in the parameter space on which we evaluate the model. We note that the position, orientation and scale of the selected slice must be chosen carefully, because it strongly affects the resulting landscapes~\cite{liVisualizingLossLandscape2018}.

\begin{figure}[htb]
    \centering
    \includegraphics[width=\linewidth]{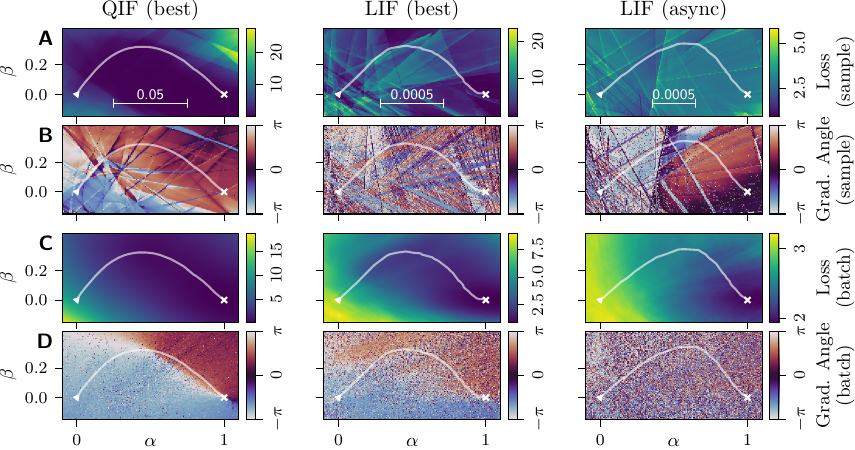}
    \caption{Loss and gradient landscapes for three selected models. (A) Sample loss landscapes for an example input sample. (B) Gradient angle landscape corresponding to the sample loss shown in A. (C) Batch loss landscapes for an example input batch containing 1{,}000 samples. (D) Gradient angle landscape corresponding to the batch loss shown in C. White lines indicate training trajectories over 200 epochs (starting at triangles). Scale bars in A show distances in parameter space. }
    \label{fig:landscapes}
\end{figure}

We focus on the region relevant for training and use a modified version of the slice selection proposed in \cite{liVisualizingLossLandscape2018}.
Specifically, we select the primary direction $u$ of the plane to be the unit vector pointing from the initial parameters $\theta_0$ to the final parameters $\theta_K$, where $K$ is the number of update steps. For the secondary direction $v$, we select the first principal component of the vectors $[\theta_1-\theta_0,\dots,\theta_{K-1}-\theta_0]$ projected onto the subspace orthogonal to $u$ (see \cref{sec:landscape_calulcation} for details).
We parametrize the points on the plane using two coordinates $(\alpha, \beta)$, where $\alpha$ varies along $u$ and $\beta$ varies along $v$. The parameters at $(\alpha, \beta)$ are then given by
\begin{align}
    \theta(\alpha, \beta) = \theta_0 + \alpha s u + \beta s v,
\end{align}
where $s = |\theta_K - \theta_0|$ is a scaling factor. Thus, unlike the approach in \cite{liVisualizingLossLandscape2018}, both the initial parameters $\theta_0$ and the final parameters $\theta_K$ lie on the plane, at coordinates $(0, 0)$ and $(1, 0)$, respectively. 

We start by taking on a large-scale perspective on the entire training trajectory (\cref{fig:overview}A,C and \cref{fig:landscapes}, \cref{fig:supp_landscapes} shows additional examples and gradient amplitude landscapes).
For all three models, the loss landscapes for a single input sample appear fragmented with gradually changing loss within and sharp loss changes on the boundaries of fragments (\cref{fig:landscapes}A).
We also observe many coinciding fragment boundaries in the loss and the angle of the loss gradients $(\pdv{L}{\alpha},\pdv{L}{\beta})$ for the same sample (\cref{fig:landscapes}B). For QIF neurons, we again observe gradually changing angles within and sharp changes on the boundaries of fragments. For LIF neurons, however, it appears as if the fragmentation is overlaid with noise, leading to erratically changing gradient directions also within fragments.

Since batches rather than individual samples are used for training, we also examine batch loss and batch gradient angle landscapes. Given the large batch size,
we expect the resulting landscapes to be qualitatively independent of the specific batch. One may assume that averaging over input samples, which averages their landscapes, results in a smoother landscape. Indeed, the batch loss landscapes visually change gradually and show no clear fragmentation (\cref{fig:landscapes}C). However, while the batch gradient landscapes also visually do not shatter into clearly delineated macroscopic fragments anymore, they appear much noisier (\cref{fig:landscapes}D and \cref{fig:supp_landscapes}). The degree of apparent noise is much larger for LIF compared to QIF neurons, with barely any macroscopic consistency in the gradient angle visible anymore for LIF neurons with asynchronous output spikes.

Taken together the loss and gradient landscapes appear notably smoother for QIF than LIF neurons, which is consistent with the performance differences between the three models.

\subsection{Discontinuities in loss and gradient}
\label{sec:discontinuities}

To examine the cause of the fragmentation of the loss landscape and the, on a large-scale, seemingly randomly changing gradients, we next visualize the landscapes on a much finer scale. 

Zooming into the sample loss and gradient landscapes reveals that the fragments visible in \cref{fig:landscapes}A in fact consist of many more, smaller fragments (\cref{fig:overview}B,D and \cref{fig:supp_landscapes_zoom}). For LIF neurons, the loss fragments are separated by discontinuous jumps, with each fragment resembling a plateau with slowly varying loss. The gradient angles are fragmented even further into a similar plateau-like structure. 

\begin{wrapfigure}{R}{0.5\textwidth}
    \centering
    \includegraphics[width=0.5\textwidth]{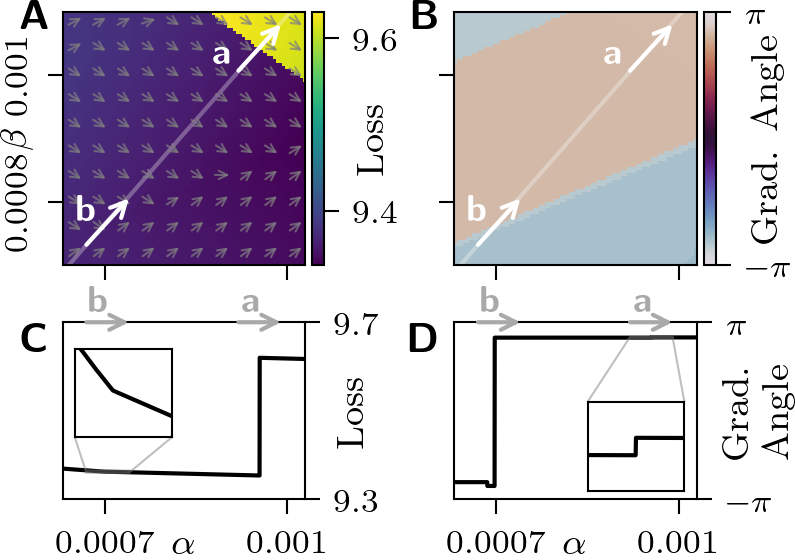}
    \caption{Example loss and gradient discontinuities in LIF neurons. (A) Sample loss landscape zoomed into a region containing both a loss discontinuity (arrow a) and a (pure) gradient discontinuity (arrow b). Gray arrows indicate direction of negative gradient. White line indicates training trajectory. (B) Gradient angle landscape of sample loss shown in (A). (C, D) Sample loss and gradient angle along the training trajectory in the same region of $\alpha$ coordinates as in (A, B).}
    \label{fig:jumps}
\end{wrapfigure}

This fine-grained fragmentation and the existence of discontinuities in both loss and gradient becomes particularly clear when focusing on a small part of a single update step within the training trajectory (\cref{fig:jumps}). The depicted example includes a discontinuous jump in the loss (arrow a in \cref{fig:jumps}), which is accompanied by a discontinuous, albeit small, jump in the gradient direction (inset in \cref{fig:jumps}D). By examining the accompanying change in network dynamics, we find that it is the result of the disappearance of a spike of a hidden-layer neuron at one point in time and the appearance of a new spike of that neuron at an earlier point in time (\cref{fig:overview}F). This (dis)appearance event is caused by a small change in the input weights, which results in the membrane potential reaching the threshold directly before the next, inhibitory (virtual) input spike arrives, which previously prevented the new output spike. In other words, the spike (dis)appearance occurs when an input and output spike time change their temporal order, as predicted from theory~\cite{klosSmoothExactGradient2025}.

The depicted example also includes a discontinuity in the gradient direction that is not accompanied by a discontinuity in the loss or any spike times (arrow b in \cref{fig:jumps}). It results from a change in how a hidden neuron reaches its threshold. Before the discontinuity, the threshold is crossed due to a virtual input spike, which immediately pushes the membrane potential above the threshold, leading to a locally constant spike time. Afterwards, due to the parameter changes, the neuron already reaches the threshold by itself because of its suprathreshold input drive, leading to a spike time with nonzero gradient (\cref{sec:lif_grad_jump}).
We note that this mechanism is only present for inputs causing voltage jumps. However, gradient discontinuities generically always occur when spikes (dis)appear, which is also the case for LIF neurons with exponentially decaying input currents, i.e.\ for inputs causing only input current jumps~\cite{klosSmoothExactGradient2025}.

In contrast and consistent with theoretical results~\cite{klosSmoothExactGradient2025}, for QIF neurons no discontinuities in the loss are present. The sharp boundaries observable in \cref{fig:landscapes}A are, instead, continuous transitions with large gradient amplitudes (\cref{sec:qif_loss_jump}). As for LIF neurons, they also agree with changes in the temporal order of input and output spikes. Yet, such events do not result in spike (dis)appearances since the effect of input spikes vanishes in the vicinity of an output spike as a result of the divergence of the membrane potential. The gradient, however, does exhibit discontinuities in such events. The mechanism underlying the gradient discontinuity described in the previous paragraph does not apply to QIF neurons, because input spikes cannot force an immediate output spike. This may be a reason why the fragmentation of the gradient landscape is not as severe as for LIF neurons.

In summary, the fragmentation of the loss and gradient landscapes results from changes in the temporal order of spikes, which cause discontinuities in both loss and gradient for LIF neurons and discontinuities in the gradient and sharp changes in the loss for QIF neurons. Further, the fragmentation is more severe for LIF neurons. The noise-like appearance of some of the landscapes on a large scale results from the inability to resolve this fine-grained fragmentation.

\subsection{Effect of discontinuities on training}
\label{sec:effects_of_disc}

\begin{figure}[htb]
    \centering
    \includegraphics[width=\linewidth]{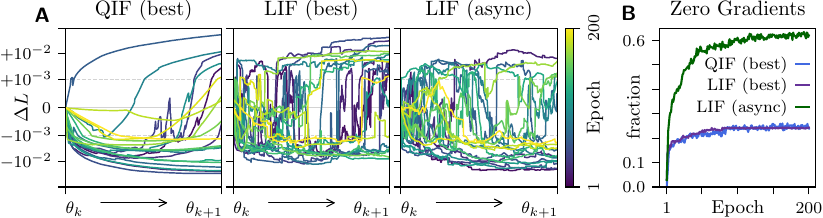}
    \caption{(A) Loss change along linear path between parameters $\theta_k$ and $\theta_{k+1}$ corresponding to the start and end of training step $k$. To reduce clutter, only the first update step of every 10-th epoch is shown (20 steps in total), each in a different color. The loss change $\Delta L(\theta) = L(\theta) - L(\theta_k)$ is shown using a symmetric log scale with a linear region between $\pm 10^{-3}$, indicated by dashed gray lines. Negative $\Delta L$ indicate a performance improvement and are thus more common in well-performing models. (B) Fraction of gradient components that vanish for all samples within each epoch. }
    \label{fig:fig5}
\end{figure}

The results of the landscape analysis suggest the existence of loss discontinuities and the increased fragmentation as reasons for the worse performance of LIF neurons. To strengthen this link, we next follow the training for one realization of the three selected models along their actual trajectories through the high-dimensional parameter space, which differ from their projection on the two-dimensional planes. Specifically, we track how the loss of the batch used in training step $k$ changes along the linear path between the parameters $\theta_k$ and $\theta_{k+1}$ before and after the update (\cref{fig:fig5}A). We observe that the loss changes much more gradually for QIF than for LIF neurons. As observed in the loss landscapes, there are further many discontinuities for the LIF neurons, in particular in the asynchronous model. 
Consistent with these observations, we find that for the QIF network 90\% of the training steps for which a loss decrease is expected from the gradient in fact results in a decrease, while this fraction is only 75\% for the best-performing LIF network and 62\% for the asynchronous LIF network (\cref{sec:dL_along_path}).
Thus, the gradient is more informative for training in QIF networks.

Finally, motivated by our observation of input spikes to LIF neurons that do not contribute to the gradient, we track the fraction of gradient components that are zero for all samples
within an epoch (\cref{fig:fig5}B). 
Larger fractions imply fewer active parameters and hence a potentially reduced expressivity. 
We find that this fraction reaches about \qty{62}{\%} for LIF neurons with asynchronous output spikes, but only about \qty{25}{\%} for both the best-performing LIF and QIF models. 
These results suggest that reduced expressivity may explain the poor performance of LIF neurons with asynchronous output spikes, but not the remaining performance gap between the best-performing LIF and QIF models.

\WFclear
\section{Discussion}
\label{sec:discussion}

We have shown that networks of QIF neurons outperform networks of LIF neurons on the SHD dataset and visualized their loss and gradient landscapes. We have found that the landscapes are strikingly fragmented, with boundaries agreeing with changes in the temporal order of spikes. For LIF networks, the fragmentation is more severe and fragments are separated by discontinuities for both the loss and gradient. For QIF networks, the fragment boundaries are only discontinuities in the gradient but not in the loss, where they are sharp but continuous. Consistent with the performance difference, this leads to a more rugged loss along the training path for LIF compared to QIF networks. Thus, the guaranteed continuity of spiking dynamics in QIF networks appears to translate into an advantage in practice.

We have focused on a single setting, i.e.\ in particular a single task and output coding scheme. Further, we have only considered intrinsically oscillating neurons and input currents that cause jumps directly in the membrane potential and not the more popular exponentially decaying input currents. This may seem restrictive, but has allowed us to perform a thorough hyperparameter search for both QIF and LIF neurons, which we deem necessary for a fair comparison, and to analyze the loss and gradient landscapes in detail. Previous work has achieved competitive performance with LIF neurons using spike-based gradient descent in different settings. However, this often requires modifying the loss function~\cite{nowotnyLossShapingEnhances2025} or using careful weight initialization and ad-hoc methods to counteract silent neurons~\cite{Eshraghian2023}. 
Together with the conceptual advantage of guaranteed continuity, our empirical results indicate that QIF neurons are more robust to the specifics of the considered setup. Nevertheless, an important task for future research is to test if our results extend to other settings as well. It would also be interesting to test if QIF neurons are advantageous when using surrogate instead of spike-based gradient descent.

In conclusion, our work suggests to replace the ubiquitously used LIF neurons with neurons exhibiting continuous spiking dynamics, such as QIF neurons, when training SNNs with gradient descent.

\begin{ack}
We thank the German Federal Ministry of Education and Research (BMBF) for support via the Bernstein Network (Bernsteinpreis 2014, 01GQ1710) and the German Research Foundation (Deutsche Forschungsgemeinschaft, DFG) for support via a Research Grant (Sachbeihilfe 557450678) and the Walter Benjamin program (Stipendium 580311533). Further we gratefully acknowledge the access to the Marvin cluster of the University of Bonn and the support provided by the HPC@HRZ Team of the University of Bonn.
\end{ack}

\bibliography{refs_standardized}

\newpage
\appendix

\input{supplementary}

\end{document}

%% file: supplementary.tex
\section{Model details}
\label{sec:model_details}

The model description was intentionally kept short in the main text to keep the focus on the main results. For completeness, we give more details about the models used in this work in this section. This is intended to be an extension of \cref{sec:models}.

\subsection{Phase representation}
\label{sec:phase_rep}
To simplify and unify the dynamics of oscillating LIF and QIF neuron models, we transform the membrane potential $V(t)$ into a phase variable $\phi(t)$ using a phase transformation $\Phi(V) = \phi$. We require the phase variable to increase linearly with time between spikes, according to the simple dynamical equation $\dot \phi = 1$, and to be reset to $\phi_\text{reset} = 0$ after reaching the threshold phase $\phi_\Theta = \lim_{V \to V_\Theta} \Phi(V)$. The following phase transformations fulfill these properties,
\begin{align}
    \Phi^\text{LIF}(V) = -\tau \log \left( 1 - \frac{V}{I_0} \right), \qquad
    \Phi^\text{QIF}(V) = \frac{\tau}{a} \left[ \arctan\left( \frac{V - \frac12}{a} \right) + \frac{\pi}{2} \right],
\end{align}
where $a = \sqrt{I_0 - \frac{1}{4}}$. The resulting threshold phases are
\begin{align}
    \phi_\Theta^\text{LIF} = -\tau \log \left( 1 - \frac{V_\Theta}{I_0} \right), \qquad
    \phi_\Theta^\text{QIF} = \frac{\pi \tau}{a}.
\end{align}
The inverse phase transformations are
\begin{align}
    (\Phi^\text{LIF})^{-1}(\phi) = I_0 \left( 1-e^{-\phi/\tau} \right), \qquad
    (\Phi^\text{QIF})^{-1}(\phi) = \frac{1}{2} + a \tan \left( \frac{a \phi}{\tau} - \frac{\pi}{2} \right).
\end{align}

For infinitesimally short input currents, the membrane potential of neuron $i$ jumps by $w_{ij}$ when receiving an input spike from neuron $j$.
The effect of this potential jump on the phase variable can be quantified using the phase transfer curve
\begin{align}
    H_w(\phi) = \Phi(\Phi^{-1}(\phi) + w).
\end{align}
The phase transfer curves of LIF and QIF neurons are
\begin{align}
    H_w^\text{LIF}(\phi) = -\tau \log\left( e^{-\phi / \tau} - \frac{w}{I_0} \right), \quad
    H_w^\text{QIF}(\phi) = \frac{\tau}{a} \left\{ \arctan \left[ \tan \left( \frac{a \phi}{\tau} - \frac{\pi}{2} \right) + \frac{w}{a} \right] + \frac{\pi}{2} \right\}.
\end{align}

Although we never show phase values directly in the main text, every network used in this work is simulated in phase representation. To reduce complexity and increase accessibility of the main text, we convert all phases into membrane potentials using the inverse transformations $\Phi^{-1}$.

\subsection{Immediate firing}
Since LIF neurons have a finite threshold potential $V_\Theta$, they can immediately fire when receiving an input spike from a neuron with a sufficiently large synaptic weight. The membrane potential of neuron $i$ immediately after receiving an input from neuron $j$ thus either jumps by $w_{ij}$ or is reset to the reset potential $V_\text{reset} = 0$. Specifically, the potential right after the input spike arrived is given by
\begin{align}
    V_i^+ = \begin{cases}
        V_i^- + w_{ij} & \text{if } V_i^- + w_{ij} < V_\Theta \\
        0 & \text{if }  V_i^- + w_{ij} \geq V_\Theta
    \end{cases},
\end{align}
where $V_i^-$ is the potential right before the input spike arrives. In phase representation (\cref{sec:phase_rep}), we have
\begin{align}
    \phi_i^+ = \begin{cases}
        H_{w_{ij}}(\phi_i^-) & \text{if } H_{w_{ij}}(\phi_i^-) < \phi_\Theta \\
        0 & \text{if } H_{w_{ij}}(\phi_i^-) \geq \phi_\Theta
    \end{cases}.
\end{align}
Note that if an LIF neuron fires immediately after an input spike, the resulting phase $\phi_i^+$ (or membrane potential $V_i^+$) does not depend on the synaptic weight $w_{ij}$. If neuron $i$ does not receive any other input spikes from neuron $j$ before it has reached the threshold, the resulting spike time of neuron $i$ does not depend on $w_{ij}$, leading to a vanishing gradient. QIF neurons do not have this problem since they have an infinite threshold potential, which can never be reached immediately due to an input spike.

\subsection{Pseudodynamics}

To enable the addition of new spikes via spike-based gradient descent, we extend a trial using so-called pseudodynamics, which guarantee that the neurons spike as often as necessary for the considered task~\cite{klosSmoothExactGradient2025}. 

The pseudodynamics are comprised of the following steps. First, at the beginning of the pseudodynamics (the trial end $T$), an interaction between all connected network neurons is enforced. The neurons in the first layer do not receive input from other network neurons. Hence, their voltage is equal to voltage at the end of the ordinary dynamics, i.e.\ $V_{\text{ps},i}^{(1)}(T)=V_i^{(1)}(T)$, where $V_{\text{ps},i}^{(1)}(T)$ is the voltage at the start of the pseudodynamics with the superscript indexing the layer number. For all other layers $l$, the voltage is 
\begin{align}
    V_{\text{ps},i}^{(l)}(T) = V_{i}^{(l)}(T) + \sum_j w^{(l)}_{ij} \frac{\Phi( V_{\text{ps},j}^{(l-1)}(T))}{\phi_\Theta},
\end{align}
where $w^{(l)}_{ij}$ are the weights of layer $l$. Afterwards, the neurons continue to evolve according to the same dynamics as during the ordinary dynamics (\cref{eq:LIF voltage,eq:QIF voltage,eq:Input current}), but excluding any further interaction between the neurons. The pseudynamics are continued until all spikes necessary for the task at hand are generated. In our case this is the case when all output neurons that have not spiked during the trial have spiked once during the pseudodynamics. The first pseudospike time $t_{\text{ps},i}^{(L)}$ of a neuron $i$ in the output layer $L$ is given by
\begin{align}
    t_{\text{ps},i}^{(L)} = T + \phi_\Theta - H_{w_{\text{ps},i}^{(L)}}(\phi_i^\text{out}(T)), \quad w_{\text{ps},i}^{(L)} = \sum_{j} w^{(L)}_{ij} \frac{\Phi( V_{\text{ps},j}^{(L-1)}(T))}{\phi_\Theta}.
\end{align}

The interaction between neurons was introduced in \cite{klosSmoothExactGradient2025} to ensure that for QIF neurons a spike time is continuous when it crosses the trial end, i.e.\ when it transitions from an ordinary spike to a pseudospike or vice versa~\cite{klosSmoothExactGradient2025}, and to enable the flow of gradients to all layers via output and hidden neurons that do not spike within the trial. The computation of pseudospike times is effectively the same as a single forward pass through a feed-forward ANN and does not necessitate iterating over spikes or timesteps as in event- or timestep-based simulations.

\subsection{Loss}
\label{sec:loss}
As usual for classification tasks, we use the cross entropy loss. The sample loss for an input sample $x$ with corresponding true label $l$ is
\begin{align}
    \label{eq:sample_loss}
    L_{(x, l)}(\theta) = -\log \frac{e^{\beta z_l(x, \theta)}}{\sum_j e^{\beta z_j(x, \theta)}},
\end{align}
where $z_i(x, \theta)$ is the logit of the $i$-th output neuron and $\beta$ is a scaling factor. Due to its interpretation in statistical physics, $\beta$ is often referred to as the inverse temperature. In this work, we call it the logit scale to emphasize its scaling effect on the logits $z_i$.
When training, we always evaluate the loss on a batch $B$ containing $N_B = 1000$ input samples with labels $(x, l)$. The mean loss across all batch samples is the batch loss
\begin{align}
    \label{eq:batch_loss}
    L_B(\theta) =  \frac{1}{N_B} \sum_{(x, l) \in B} L_{(x, l)}(\theta).
\end{align}
The model prediction for a single input sample $x$ is the output index $i$ that has the largest corresponding logit $z_i$. In this work, we focus on time-to-first-spike coding, which assigns larger logits to output neurons that spike earlier, according to
\begin{align}
    z_i(x, \theta) = -\frac{t_i(x, \theta)}{T}.
\end{align}
Here, $t_i(x, \theta)$ denotes the first spike time of output neuron $i$, given the input sample $x$ and the model parameters $\theta$, and $T$ is the trial duration, which serves as a normalization factor. The predicted label is thus the index of the output neuron that spikes first. 

Although pseudodynamics are useful during model training, we would like our final trained model to perform well without them during inference. On the one hand, this reduces the complexity of the final model and, on the other hand, should allow for an easier implementation on neuromorphic hardware. To make sure the model does not rely on pseudospikes, we introduce a regularization term, which punishes pseudospikes that are required for inference,
\begin{align}
    L_{(x, l)}^\text{reg}(\theta) = \eta \Big[ \min \big( z_l(x, \theta) - \tilde z, 0 \big) \Big]^2.
\end{align}
Here, $\eta = 10^3$ is the regularization strength and $\tilde z = -0.98$ is the target logit. If the logit is larger than this
target logit, it does not depend on pseudospikes and thus is not punished. The offset of $0.02$
is included to ensure that the spike time is safely inside the trial, even if small variations
occur. We square the distance to get a well-defined gradient at $z_l(x, \theta) = \tilde z$.

\subsection{Initialization}

The weights of network layer $l$ are initialized using a uniform Kaiming initialization \cite{heDelvingDeepRectifiers2015}, as was done in \cite{cramerHeidelbergSpikingData2022},
\[
    w_{ij}^{(l)} \sim \mathcal U \left( -\frac{w_0}{\sqrt{N_{l-1}}}, \frac{w_0}{\sqrt{N_{l-1}}} \right).
\]
Here, $w_0$ is a hyperparameter that can be used to scale the initial weights. The neuron phase variables are initialized at $\frac{\phi_\Theta}{2}$. We also experimented with learnable and random phase initializations but did not find a significant advantage in ad-hoc experiments.

\subsection{Event-based simulation}
\label{sec:simulation_details}

We simulate each network layer in an event-based manner, i.e., by processing each spike one-by-one and evolving the dynamics between spikes analytically. Since we only consider feedforward networks, each layer can be simulated separately. Each simulation step, indexed by $k$, starts at the previous spike time $t_k$, or at the initial time $t_0 = 0$. The phase of neuron $i$ in at the beginning of step $k$ is denoted by $\phi_{ik}$. The simulation of step $k$ then involves the following operations:
\begin{enumerate}
    \item Determine the neuron index $j_{k+1}$ and spike time $t_{k+1}$ of the next spike. This could either be an input spike or a spike in the layer that is simulated, depending on which appears earlier.
        Because the phase grows linearly with time between spikes, the neuron that spikes next is always the one with the largest phase, $j_{k+1} = \text{argmax}_i \phi_{ik}$.
    \item Evolve neurons freely until shortly before the next spike at $t_{k+1}^-$. This is trivial due to the linear dynamics between spikes. We have $\dot \phi_i = 1$ and, thus, $\phi_i(t_{k+1}^-) = \phi_{ik} + (t_{k+1} - t_k)$.
    \item If the next spike is an input spike, transfer it to the neurons in the simulated layer, using the phase transfer function $\phi_{i,k+1} = H_{w_{ij_{k+1}}}(\phi_i(t_{k+1}^-))$. Otherwise, if the spike occurs in the simulated layer itself, reset the phase of the spiking neuron, $\phi_{j+1,k+1} = 0$, and set $\phi_{i,k+1} = \phi_i(t_{k+1}^-)$ for all other neurons.
\end{enumerate}
This process has to be repeated until the simulation reaches the trial end $T$, i.e., until $t_{k+1} \geq T$.

\section{Data preprocessing and augmentation}
\label{sec:data_preprocessing}
To reduce computational complexity and to improve the generalization abilities of our models, we use several data preprocessing and augmentation strategies.

\subsection{Temporal binning}
\label{sec:temporal_binning}

The SHD samples contain between 2{,}410 and 16{,}257 spikes. To avoid iterating over all of them, which is computationally inefficient, we discretize the trial duration $T$ into $N_{\text{T}}$ bins. Bin $k$, $k=1,\dots,N_{\text{T}}$, covers the range $\frac{(k - 1) T}{N_\text{T}} \leq t_j < \frac{k T}{N_\text{T}}$. Then, we replace all input spikes within each bin $k$ by a single virtual input spike located at the bin center,
\begin{align}
    t_k^{\text{virt}}
    =
    \frac{\left(k-\frac{1}{2}\right)T}{N_{\text{T}}},
\end{align}
where $T$ is the trial duration. We choose $T=\SI{1.4}{s}$, which fully covers the SHD spike-time range (last input spike at \qty{1.369}{\second}).

This affects the input currents as follows. Without introducing virtual input spikes, the input current of neuron $i$ in the first layer is given by
\begin{align}
    I_i^{(1)}(t) &= I_0 + \tau\sum_{j=1}^{N_0}w_{ij}^{(1)}\sum_{t_j}\delta(t-t_j),
\end{align}
where $N_0$ is the number of input neurons and $w_{ij}^{(1)}$ are the input weights. We then approximate all spike times within bin $k$ by $t_k^{\text{virt}}$, which yields
\begin{align}
    I_i^{(1)}(t) &\approx I_0 + \tau\sum_{j=1}^{N_0} w_{ij}^{(1)} \sum_{k=1}^{N_{\text{T}}}n_{jk}\, \delta(t-t_k^{\text{virt}}),
\end{align}
where $n_{jk}$ denotes the number of spikes of input neuron $j$ in bin $k$. Finally, defining the virtual input weights as
\begin{align}
    w_{ik}^{\text{virt}} &= \sum_{j=1}^{N_0} w_{ij}^{(1)} n_{jk},
\end{align}
the input current becomes
\begin{align}
    I_i^{(1)}(t) &\approx I_0 + \tau \sum_{k=1}^{N_{\text{T}}} w_{ik}^{\text{virt}} \delta(t-t_k^{\text{virt}}).
\end{align}

This formulation is equivalent to replacing the original input layer by $N_{\text{T}}$ virtual neurons, each emitting a single spike at time $t_k^{\text{virt}}$. The virtual weights are not trainable parameters, but are fully determined by the learnable input weights $w_{ij}^{(1)}$ and the spike counts $n_{jk}$.

\subsection{Spatial binning}
In addition to the temporal binning, we also bin the input data along the spatial direction, which is given by the input neuron indices $j = 1, ..., N_0$. Compared to the time direction, the spatial direction is already discretized into $N_0 = 700$ bins in the SHD dataset, if we regard each neuron as a single bin. To improve the ability of our model to generalize to new data and to reduce the number of learnable parameters, we reduce the spatial resolution. To do this, we squash together multiple input neurons into one by superposing their spike trains, as was done in \cite{cramerHeidelbergSpikingData2022}. If $M_\text{S}$ is the number of neurons that are combined, i.e., the number of neurons in each spatial bin, the number of spatial bins is $N_\text{S} = N_0 / M_\text{S}$. Assuming that we already applied temporal binning before and the corresponding spike counts are $n_{jk}$, the spike count in time bin $k$ and spatial bin $m$, $m = 1, ..., N_\text{S}$, is
\[
    \tilde n_{mk} = \sum_{j = (m - 1) M_\text{S} + 1}^{m M_\text{S}} n_{jk},
\]
where we sum over all neurons in the $m$-th bin. The feedforward weight matrix of the first layer reduces from an $N_1 \times N_0$ to an $N_1 \times N_\text{S}$ matrix with elements $\tilde w_{im}^{(1)}$.
The virtual weights from the previous section then are
\[
    \tilde w_{ik}^{\text{virt}} = \sum_{m=1}^{N_\text{S}}  \tilde w_{im}^{(1)} \tilde n_{mk}.
\]

\subsection{Spatial shift}
Another method to improve the generalization performance is to randomly shift the input spikes in the spatial direction, i.e., across neuron indices. This shift is applied globally for each sample, so every spike in a single sample is shifted equally. However, in each epoch, a different shift is applied, such that the model does not always process the same shifted data, which would lead to a bias and not improve generalization at all. Since we precompute the histograms resulting from temporal and spatial binning before the training, we apply the shift to the spatial bin index $m$ instead of the neuron index itself. The global shift was also applied in \cite{nowotnyLossShapingEnhances2025} and in fact lead to better test accuracies. As done there, we draw the shift distance $\Delta m$ from a uniform distribution and truncate the decimals to get an integer value\footnote{While in \cite{nowotnyLossShapingEnhances2025} it is stated that values are rounded to the nearest integer, the implementation truncates the decimals instead (i.e., rounds toward zero). We therefore also truncate them.},
\[
    \Delta m \sim \text{trunc} \left[ \mathcal U \left(-(\Delta m_\text{max} + 1), \Delta m_\text{max} + 1 \right) \right].
\]
Here, we add one to the maximum shift distance $\Delta m_\text{max}$ since otherwise, due to the truncation, the maximum value would effectively never be drawn. The shifted bin is now computed by adding the random shift distance, $m' = m + \Delta m$. At the boundary towards which the bins are shifted, the spikes are dropped. In the histograms resulting from spatial and temporal binning, they do thus not contribute. At the opposite boundary, neurons without any spikes remain, represented by zeros in the histogram.

\subsection{Delayed copies}
As done in \cite{nowotnyLossShapingEnhances2025}, we augment the input data by adding $N_\text{D}$ delayed copies of each sample. The $k$-th copy is delayed by $k \cdot d$, where $d$ is the delay between two copies. Spikes that end up occurring after the trial end $T$ are discarded. As a result, the dataset size increases by a factor of $N_\text{D} + 1$.

Using this augmentation, we intend to improve generalization performance by increasing the robustness to temporal shifts in the input data. For spoken digits, translations in time do not change the meaning behind them. Using delayed input copies, we can train the model on the same original sample, but at a different time during the trial.

\section{Hyperparameter search}
\label{sec:hparam_search}

\begin{figure}[p]
    \centering
    \includegraphics[width=\linewidth]{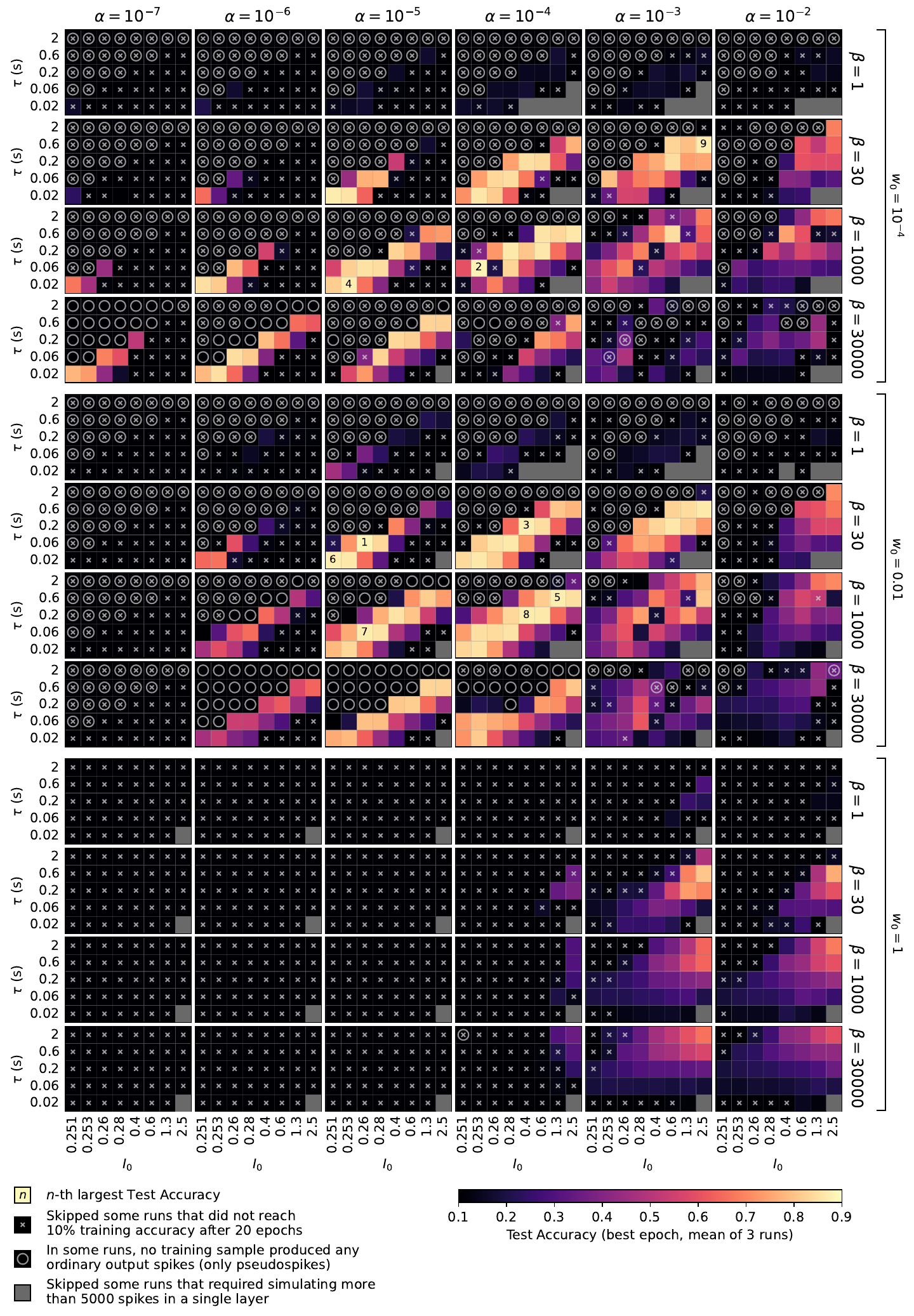}
    \caption{Results of rough hyperparameter search with QIF neurons. For each configuration, we show the best test accuracy achieved across all 100 epochs, averaged over 3 random seeds.}
    \label{fig:scan_QIF}
\end{figure}

\begin{figure}[p]
    \centering
    \includegraphics[width=\linewidth]{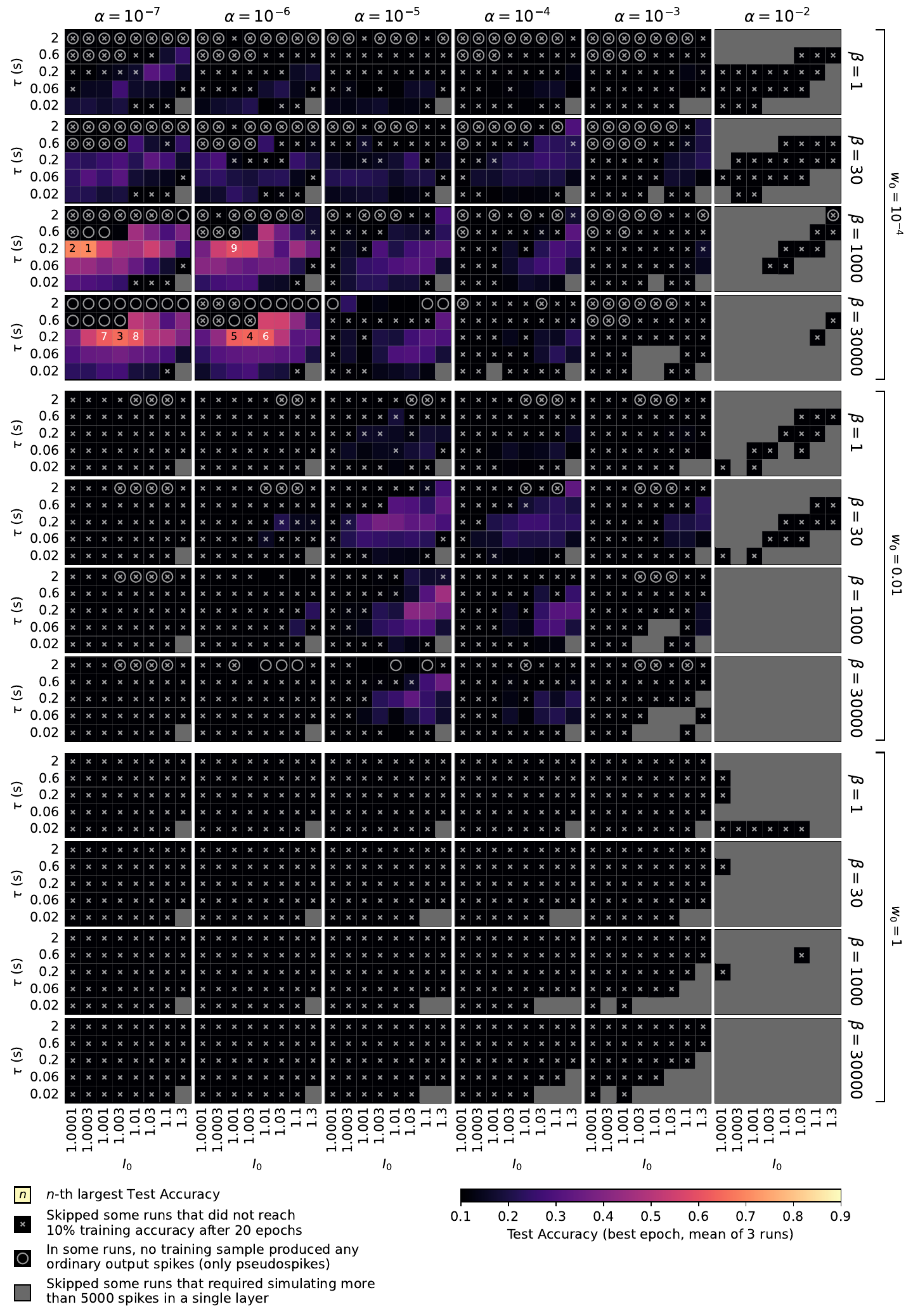}
    \caption{Results of rough hyperparameter search with LIF neurons. For each configuration, we show the best test accuracy achieved across all 100 epochs, averaged over 3 random seeds.}
    \label{fig:scan_LIF}
\end{figure}

First, we perform a rough hyperparameter search, varying the weight scale $w_0$,  the learning rate $\alpha$, the logit scale $\beta$, the membrane time constant $\tau$, and the constant input drive $I_0$ over a wide range of values. For each hyperparameter configuration, we simulate 100 epochs with three different random seeds and determine the best test accuracy the model achieves during training in each run. In \cref{fig:scan_QIF,fig:scan_LIF}, we show these best test accuracies for networks of QIF and LIF neurons, respectively. It should be noted that using the test dataset for determining the best hyperparameter configuration is bad practice since this might cause overfitting on the test set and an overestimation of the accuracy on new data. In this work, we still use the test set for model selection for a fair comparison with other works, which do the same (e.g., see \cite{cramerHeidelbergSpikingData2022,bittarSurrogateGradientSpiking2022,schoneScalableEventbyeventProcessing2024a}). A better approach would be to perform cross-validation, as was done, e.g., in \cite{nowotnyLossShapingEnhances2025}. However, this would introduce additional computational and conceptual complexity, which we avoid here due to our focus on the detailed analysis of the models themselves.

\begin{figure}[p]
    \centering
    \includegraphics[width=0.95\linewidth]{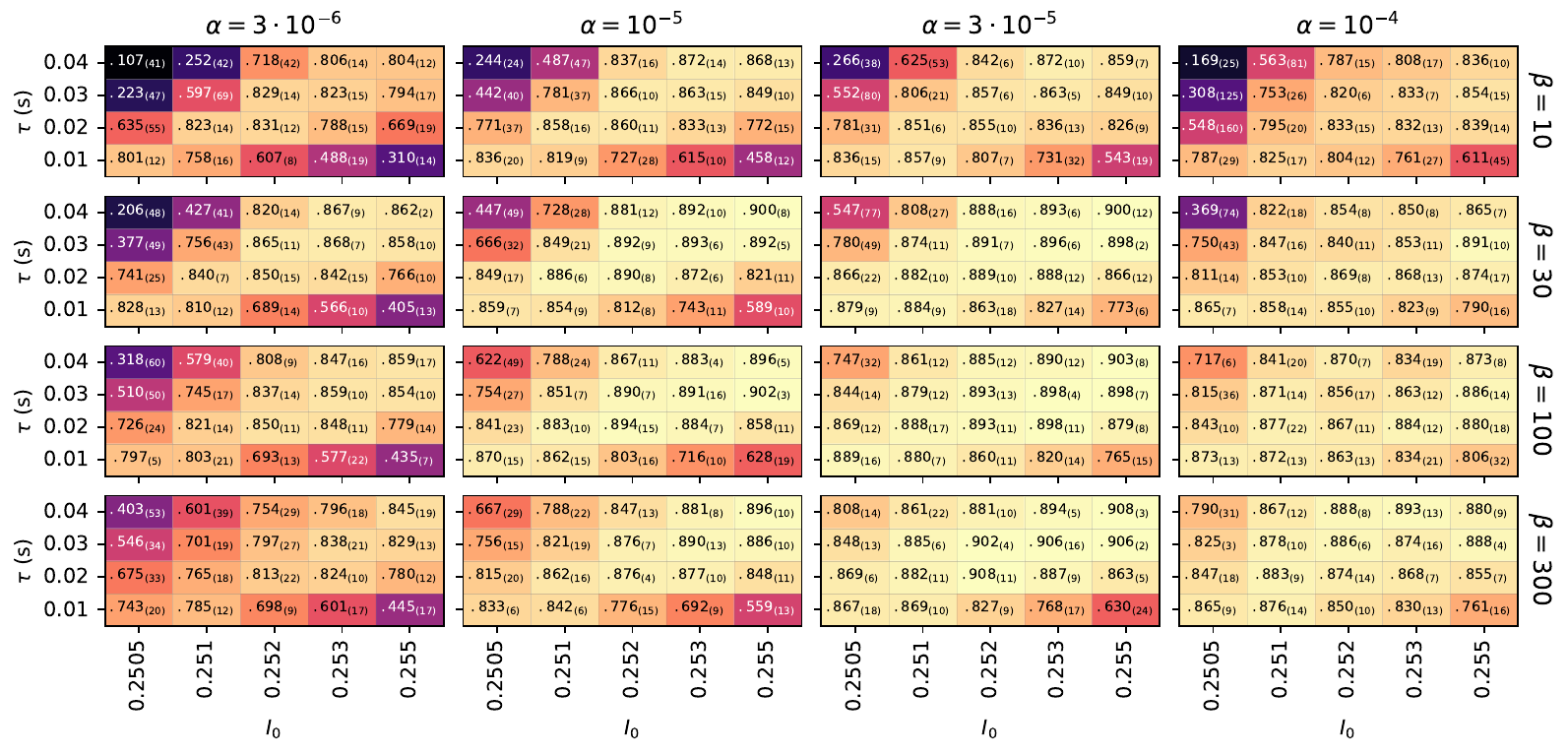}
    \caption{Results of fine hyperparameter search with QIF neurons. For each configuration, we show the best accuracy achieved across all 200 epochs, averaged over 5 random seeds.}
    \label{fig:scan_fine_QIF}
\end{figure}

Next, we focus on the well-performing regions in a finer hyperparameter search, covering smaller volumes in the hyperparameter space in more detail and training for 200 instead of only 100 epochs. For the QIF model, we fix $w_0 = 0.01$ and scan a portion of the region with high accuracies in \cref{fig:scan_QIF} with a finer resolution. Note that there is a degree of freedom for the selection of the exact position of the finer search because of the invariance of QIF neurons with respect to a simultaneous scaling of $\tau$, $\sqrt{I_0-1/4}$ and $w_0$ (cf.\ \cref{sec:model_details}). The resulting test accuracies are shown in \cref{fig:scan_fine_QIF}. The highest test accuracy we find is $(90.8 \pm 0.3)\%$ at $\alpha = 3 \cdot 10^{-5}$, $\beta = 300$, $\tau = \SI{0.04}{s}$, and $I_0 = 0.252$.

\begin{figure}[p]
    \centering
    \includegraphics[width=0.95\linewidth]{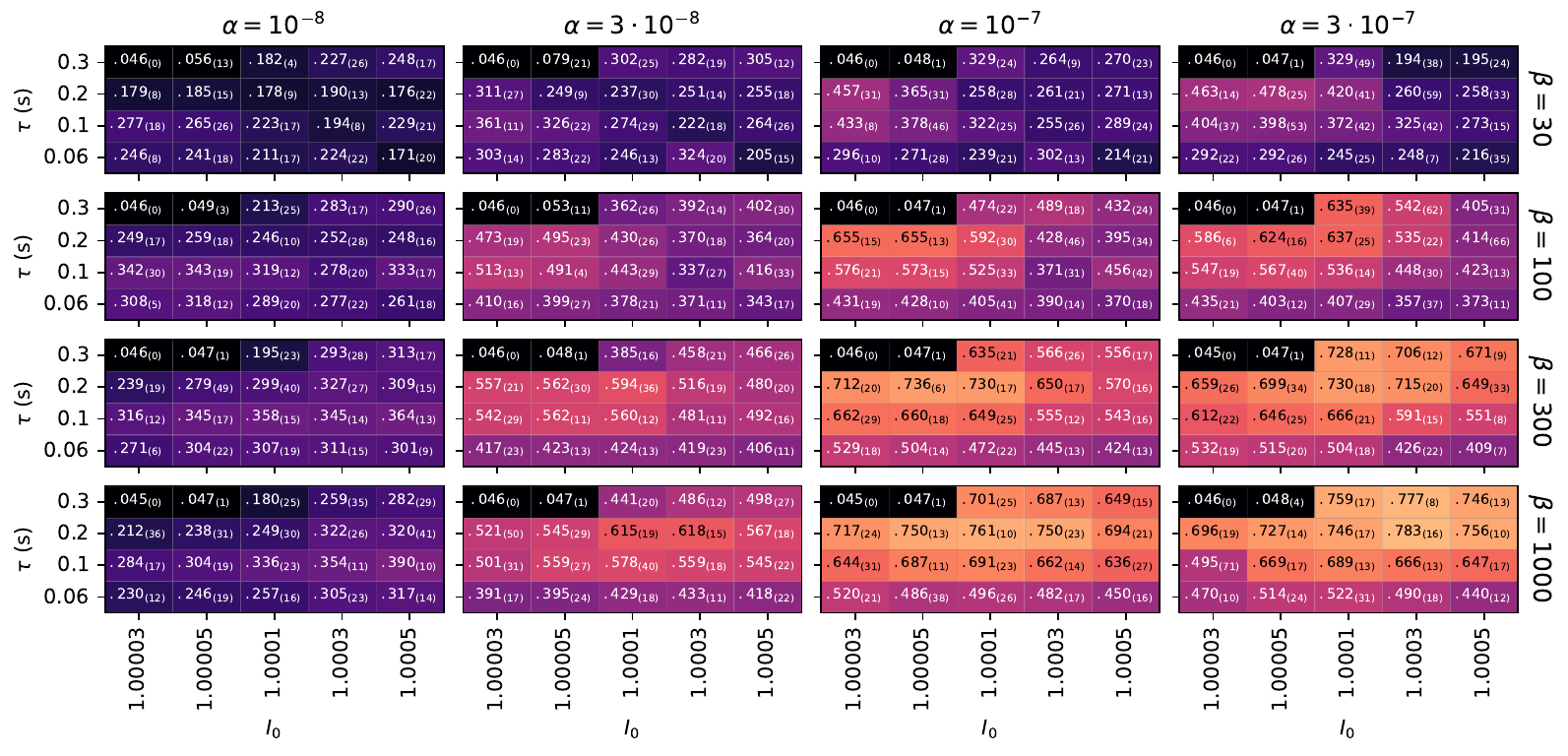}
    \caption{Results of fine hyperparameter search with LIF neurons and low $I_0$. For each configuration, we show the best accuracy achieved across all 200 epochs, averaged over 5 random seeds.}
    \label{fig:scan_fine_LIF_low_I0}
\end{figure}

\begin{figure}[p]
    \centering
    \includegraphics[width=0.95\linewidth]{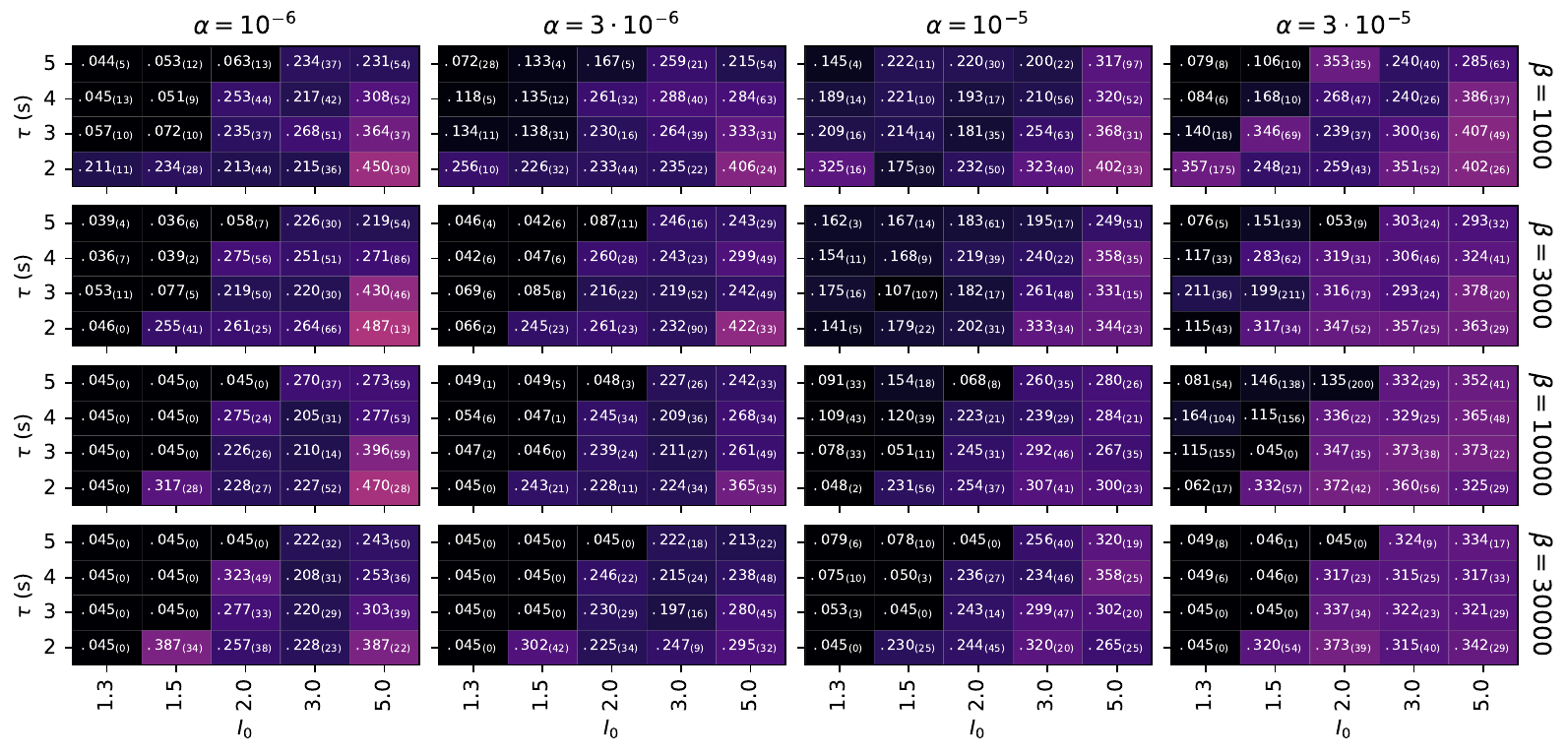}
    \caption{Results of fine hyperparameter search with LIF neurons and high $I_0$. For each configuration, we show the best accuracy achieved across all 200 epochs, averaged over 5 random seeds.}
    \label{fig:scan_fine_LIF_high_I0}
\end{figure}

For the LIF neuron, we fix $w_0 = 10^{-4}$ and perform two separate fine hyperparameter searches, one for low $I_0$ and $\tau$ around \SI{0.2}{s}, and one for high $I_0$ and higher $\tau$, since those are the regions with high test accuracies in \cref{fig:scan_LIF}. In both searches, we extend the search space beyond the boundary of the rough search. The resulting test accuracies are shown in \cref{fig:scan_fine_LIF_low_I0,fig:scan_fine_LIF_high_I0}. We see that the region with lower $I_0$ and lower $\tau$ performs significantly better than the other region, achieving up to $(78.3 \pm 1.6) \%$ test accuracy.  \cref{tab:hparams} shows the final hyperparameters and metrics for both QIF and LIF networks.


\begin{table}[htb]
    \centering
    \begin{tabular}{lccc}
        \toprule
        & QIF (best) & LIF (best) & LIF (async)  \\
        \midrule
        Initial weight scale $w_0$ & $10^{-2}$ & $10^{-4}$ & $10^{-4}$ \\
        Learning rate $\alpha$ & $3 \cdot 10^{-5}$ & $3 \cdot 10^{-7}$ & $10^{-7}$ \\
        Logit scale $\beta$ & 300 & 1{,}000 & 100 \\
        Membrane time constant $\tau$ & \qty{20}{ms} & \qty{200}{ms} & \qty{200}{ms} \\
        Constant input drive $I_0$ & 0.252 & 1.0003 & 1.00003 \\
        \midrule
        Layer sizes & \multicolumn{3}{c}{(128, 20)} \\ 
        Trial duration $T$ & \multicolumn{3}{c}{ \qty{1.4}{s} } \\
        Batch size $N_B$ & \multicolumn{3}{c}{  1{,}000 } \\
        Number of spatial bins $N_S$ & \multicolumn{3}{c}{  35 ($\rightarrow$ 20 neurons per bin) } \\
        Maximum spatial shift $\Delta m_\text{max}$ & \multicolumn{3}{c}{  2 bins ($\hat{=}\: 40$ neurons) } \\
        Number of temporal bins $N_T$ & \multicolumn{3}{c}{  64 ($\rightarrow$ bin size \qty{21.875}{ms}) } \\
        Number of delayed input copies $N_D$ & \multicolumn{3}{c}{  2 } \\
        Delay per input copy $d$ & \multicolumn{3}{c}{ \qty{0.1}{s} } \\
        \midrule
        Distinctness (final, test set) & \qty{28.3(14)}{ms} & \qty{3.12(9)}{ms} & \qty{29.2(46)}{ms} \\
        Training accuracy (final) & \qty{97.0(5)}{\%} & \qty{83.5(11)}{\%} & \qty{45.6(18)}{\%} \\
        Test accuracy (final) & \qty{90.1(9)}{\%} & \qty{79.2(13)}{\%} & \qty{50.8(40)}{\%} \\
        \bottomrule
    \end{tabular}
    \vspace{4pt}
    \caption{Hyperparameters of the three models selected for further analysis and the corresponding metrics after training for 500 epochs. For the final accuracy and distinctness values, we report mean and standard deviation over ten seeds.}
    \label{tab:hparams}
\end{table}

\section{Large network results}
\label{sec:large_network_results}
To achieve higher accuracies than in \cref{tab:hparams}, we use larger networks and perform additional exploratory experiments, also varying hyperparameters that were not part of the hyperparameter searches. \cref{tab:large_network_hparams} shows the hyperparameters and metrics of the best large models, both trained for 500 epochs and 9 different random seeds. Especially the LIF network profits from the increased number of neurons. However, we did not find that multiple hidden layers improve the LIF performance. For QIF networks, we found the best accuracy using two hidden layers. Note that in intermediate epochs, the same QIF and LIF networks achieved slightly higher test accuracies of \qty{93.8(6)}{\%} in epoch \qty{162(25)}{} and \qty{88.2(8)}{\%} in epoch \qty{452(43)}{}, respectively, averaged over 9 seeds. This indicates that early stopping or reducing overtraining might be beneficial, but we did not look into this in more detail since accuracy optimization is not the main focus of this work.
\begin{table}[htb]
    \centering
    \begin{tabular}{lcc}
        \toprule
        & QIF (large) & LIF (large)  \\
        \midrule
        Layer sizes & (512, 512, 20) & (1024, 20) \\
        Logit scale $\beta$ & 100 & 1{,}000 \\
        \midrule
        Distinctness (final, test set) & \qty{20.5(5)}{ms} & \qty{5.65(20)}{ms} \\
        Training accuracy (final) & \qty{99.973(12)}{\%} & \qty{94.08(41)}{\%}\\
        Test accuracy (final) & \qty{91.3(8)}{\%} & \qty{85.2(13)}{\%}  \\
        \bottomrule
    \end{tabular}
    \vspace{4pt}
    \caption{Hyperparameters and metrics of the two best-effort large QIF and LIF networks, if different from QIF (best) and LIF (best) in \cref{tab:hparams}. Metrics are averaged over 9 runs with different seeds.}
    \label{tab:large_network_hparams}
\end{table}

\section{Accuracy-distinctness relation}
\label{sec:acc_vs_distinctness}

\begin{figure}[h!]
    \centering
    \includegraphics[width=\linewidth]{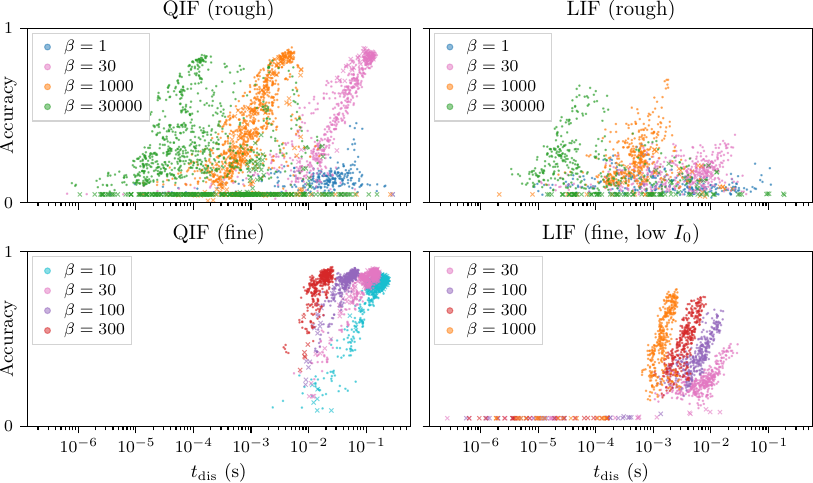}
    \caption{Final test accuracy plotted against distinctness for hyperparameter configurations on the rough and fine search grid, grouped by the neuron type and colored by the logit scale $\beta$. To reduce clutter, we exclude the fine LIF grid search with high $I_0$. Runs with only pseudospikes in the output layer are indicated by crosses instead of dots.}
    \label{fig:acc_vs_distinctness}
\end{figure}

In \cref{fig:model_comparison}E, we show how the distinctness compares to the final test accuracy for the configurations of the rough hyperparameter search. Here, we look at the accuracy-distinctness relation for both hyperparameter searches in a more detail. They are shown in \cref{fig:acc_vs_distinctness}. We notice that the distinctness is highly dependent on the logit scale $\beta$, which is indicated by the color of the points. Configurations with larger logit scales usually have smaller distinctness values. This is likely because of the form of the sample loss function in \cref{eq:sample_loss}, larger logit scales require smaller logit changes to achieve the same change in loss. Thus, the output spikes are not spread out as much over time during training, reducing the distinctness. Comparing the QIF and LIF plots, it becomes apparent that QIF neurons allow much larger distinctness values without sacrificing accuracy while LIF neurons suffer from an accuracy reduction when requiring a higher distinctness. However, since hyperparameter searches are necessarily restricted to finite regions of the parameter space, a complete characterization of the underlying landscape cannot be guaranteed.

\section{Landscape calculation}
\label{sec:landscape_calulcation}

Here, we give a more detailed description of the landscape construction, introduced in \cref{sec:landscapes}.
Our goal is to evaluate a function $f(\theta)$, e.g.\ the loss $L(\theta)$, on a two-dimensional subspace $\mathbb R^2$ of the parameter space $\mathbb R^N$, where $N$ is the number of trainable parameters. We call the subspace the projection plane. The plane is spanned by two normalized and orthogonal direction vectors $u, v \in \mathbb R^N$. We choose the primary direction vector $u$ such that it points from the initial model parameters $\theta_0 \in \mathbb R^N$ to the final parameters $\theta_K \in \mathbb R^N$ after $K$ training steps,
\begin{align}
    u = \frac{\theta_K - \theta_0}{|\theta_K - \theta_0|}.
\end{align}
Let $\theta_k \in \mathbb R^N$ be the parameter vector after $k = 0, 1, ..., K$ training steps. 
The secondary direction vector $v$ should point in the direction orthogonal to $u$ along which the training trajectory exhibits its largest variance.
To quantify this, we consider the perpendicular distance vectors $d_k \in \mathbb R^N$ from the line through $\theta_0$ and $\theta_K$ to the parameter vectors $\theta_k$, which are given by
\begin{align}
    d_k = \theta_k - \theta_0 - ((\theta_k - \theta_0) \cdot u) u.
\end{align}
Now, we define the secondary direction vector $v$ as the first principle component (PC1) of the distance vectors $d_k$ for $k = 1, ..., K-1$, i.e., the direction of maximum variance. Here, we exclude $d_0$ and $d_K$ because they always vanish. As intended, the resulting direction vectors $u, v$ form an orthonormal basis of the projection plane since  they satisfy $|u| = |v| = 1$ and $u \cdot v = 0$. At this point, the sign of $v$ is arbitrary. To create a consistent landscape orientation, we choose the sign such that $v$ points in the direction of the first updated parameters $\theta_1$ by requiring $d_1 \cdot v > 0$.
We ignore the edge-case $d_1 \cdot v = 0$ since it is highly unlikely.

Next, we construct a coordinate system on the projection plane. Let $\alpha, \beta \in \mathbb R$ be the coordinates in the primary ($u$) and secondary ($v$) direction, respectively. Both the initial and final parameters should lie on the projection plane. We position the initial parameters $\theta_0$ at the origin of the coordinate system, i.e., at $(\alpha, \beta) = (0, 0)$, and the final parameters $\theta_K$ at $(\alpha, \beta) = (1, 0)$. Then, the parameters at $(\alpha, \beta)$ are given by
\begin{align}
    \theta(\alpha, \beta) = \theta_0 + \alpha s u + \beta s v,
\end{align}
where $s = |\theta_K - \theta_0|$ determines the scaling.
When projecting a parameter vector $\theta \in \mathbb R^N$ onto the plane, the resulting coordinates are
\begin{align}
    \alpha(\theta) = \frac{1}{s} (\theta - \theta_0) \cdot u, \quad
    \beta(\theta) = \frac{1}{s} (\theta - \theta_0) \cdot v.
\end{align}

\section{Example of an LIF gradient discontinuity}
\label{sec:lif_grad_jump}
\begin{figure}[h!]
    \centering
    \includegraphics[width=\linewidth]{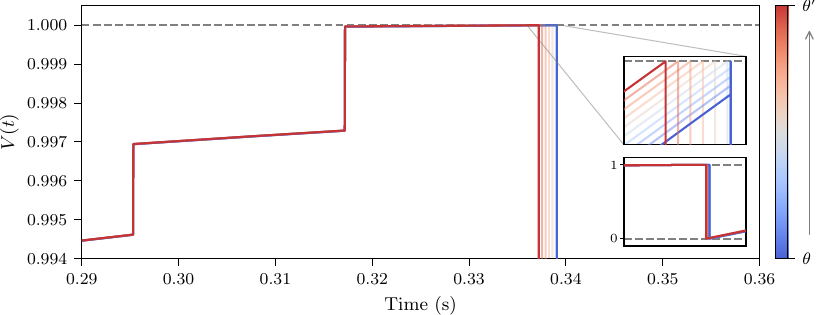}
    \caption{Change of the voltage dynamics of the hidden neuron that results in the gradient discontinuity along arrow b in \cref{fig:jumps}. Coloring as in \cref{fig:overview}F. See description of \cref{fig:jumps} in the main text for details.}
    \label{fig:arrow_b}
\end{figure}

\section{Example of a sharp QIF loss transition}
\label{sec:qif_loss_jump}
\begin{figure}[h!]
    \centering
    \includegraphics[width=\linewidth]{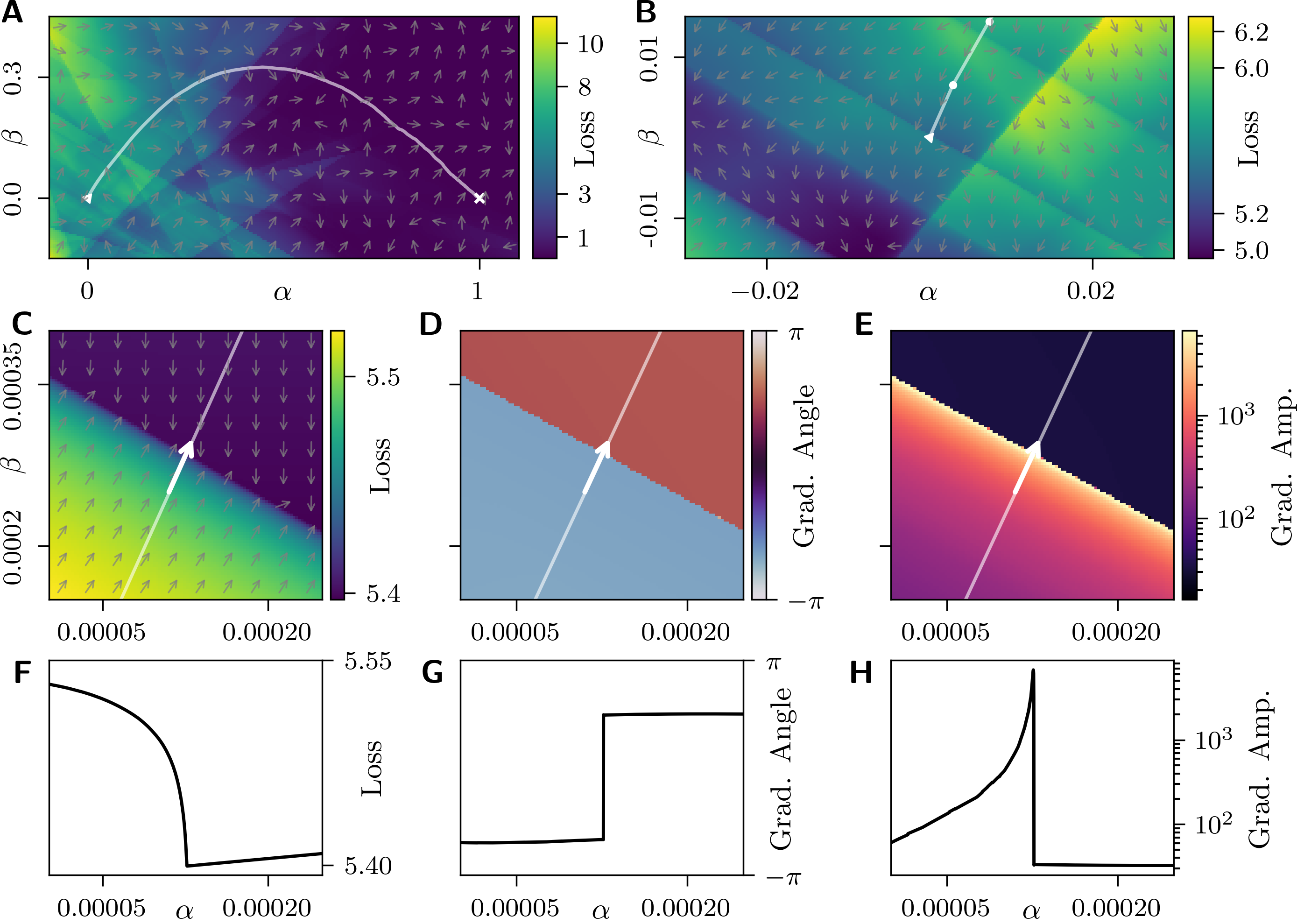}
    \caption{Example of a region of a QIF sample landscape exhibiting a sharp loss transition. (A) Large-scale view of the loss landscape including the full training trajectory (white line). (B) Fine-scale view of the loss landscape showing the beginning of the training trajectory. (C, F) Even finer scale view of the first sharp loss change along the training trajectory, demonstrating its continuity. (D, G) Like C,D, but showing the gradient angle, which exhibits a discontinuity at the sharp loss change. (E,F) Like C,D, but showing the gradient amplitude, which exhibits a discontinuity at the sharp loss change.
    }
    \label{fig:jumps_qif}
\end{figure}

Here, we present an example of a loss jump in a QIF network. In \cref{fig:jumps_qif}A, we show the loss landscape of a single sample containing the full training trajectory (200 epochs). Zooming in by a factor of 20, in \cref{fig:jumps_qif}B, we notice that there seems to be a loss jump around the initial parameters (indicated by a triangle). Zooming in even further, by a factor of 200, in \cref{fig:jumps_qif}C, we notice that the jump is a sharp continuous transition. To make this even clearer, we show the loss along the training trajectory in this region as a line plot in \cref{fig:jumps_qif}F. There, we see a very steep decline, but not a discontinuity. On the other hand, the loss gradient angle exhibits a discontinuous transition, as can be seen in \cref{fig:jumps_qif}D and G. The fact that the loss changes very rapidly in this region is also reflected by a very large loss gradient amplitude, shown in \cref{fig:jumps_qif}E and H. \cref{fig:overview}E shows an accompanying sharp but continuous change of a spike time in the network.

\section{Additional landscapes}
\label{sec:landscapes_supp}
\begin{figure}[p]
    \centering
    \includegraphics[width=\linewidth]{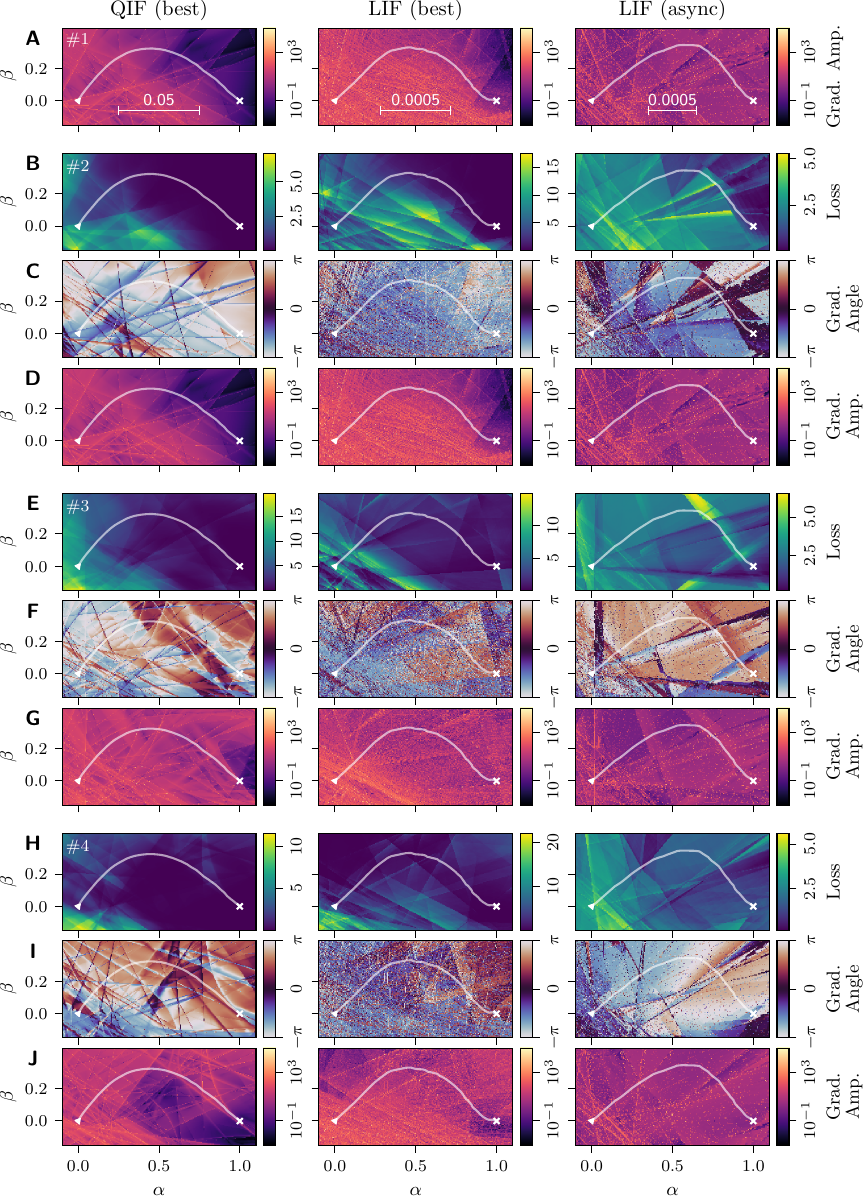}
    \caption{Additional landscapes. (A) Gradient amplitude landscapes of the same sample shown in \cref{fig:landscapes}A and B. (B-J) Loss, gradient angle and gradient amplitude landscapes for three additional samples.}
    \label{fig:supp_landscapes}
\end{figure}

\begin{figure}[p]
    \centering
    \includegraphics[width=\linewidth]{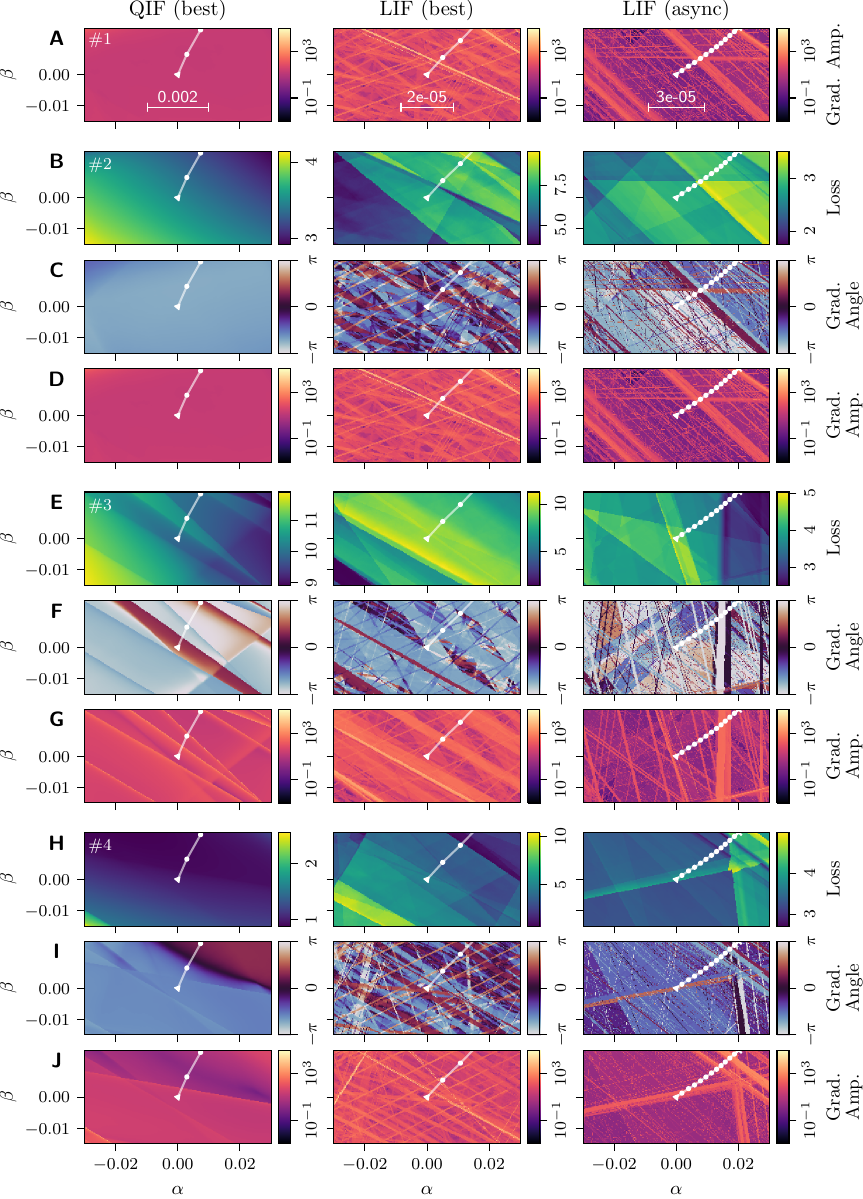}
    \caption{Same landscapes as in \cref{fig:supp_landscapes} but zoomed in by a factor of 20. Individual training steps are indicated by white circles.}
    \label{fig:supp_landscapes_zoom}
\end{figure}

In \cref{fig:landscapes}A and B, we only show the loss and gradient angle landscapes of a single sample. Here, we also show the gradient amplitude landscape of this sample in \cref{fig:supp_landscapes}A. In \cref{fig:supp_landscapes}B-J, we further show the landscapes of three additional samples. To make individual jumps more obvious, we zoom in by a factor of 20 in \cref{fig:supp_landscapes_zoom}.

\section{Loss changes along training path}
\label{sec:dL_along_path}

In \cref{fig:fig5}A, we give some examples of how the loss changes between two training steps. Some of the loss curves vary erratically, especially for LIF neurons. In some cases, the loss even increases. 
Here, we measure the loss difference $\Delta L = L(\theta_{k+1}) - L(\theta_k)$ for all training steps in 200 epochs (1{,}800 steps in total).

\begin{figure}[h!]
    \centering
    \includegraphics[width=\linewidth]{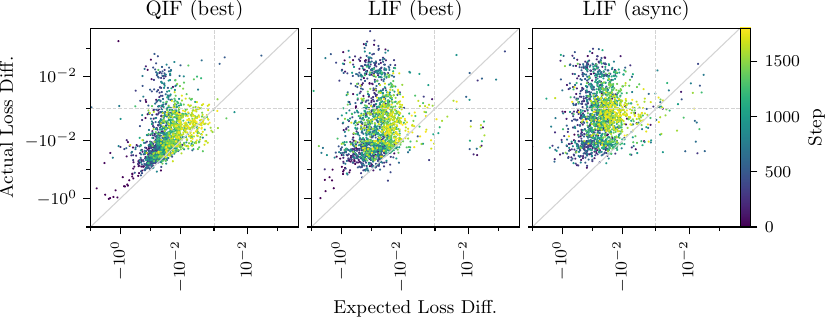}
    \caption{Actual loss differences plotted against expected ones for all training steps. Corresponding step numbers are indicated by different colors.}
    \label{fig:dL_correlation}
\end{figure}

If the loss landscape is locally well approximated by a linear expansion, the loss function satisfies
\begin{align}
    L(\theta_k + \Delta \theta_k) \approx L(\theta_k) + \Delta \theta_k \cdot \nabla L(\theta_k)
\end{align}
for sufficiently small updates $\Delta \theta_k = \theta_{k+1}-\theta_k$. Based on this first-order approximation, we define the expected loss difference as
\begin{align}
    \Delta L_\text{exp}(\theta_k) = \Delta \theta_k \cdot \nabla L(\theta_k).
\end{align}
Comparing the actual and expected loss differences therefore provides a measure of how accurately the first-order approximation describes the loss along the training trajectory (\cref{fig:dL_correlation}). 
Note that there are steps in all models with a positive expected loss difference, which result from using the Adam optimizer, which does not strictly follow the direction of steepest descent.

Gradient descent relies on a at least roughly accurate first-order approximation. As a quantitative measure of it, we determine the fraction of training steps that reduce the loss ($\Delta L < 0$), when a reduction in loss is expected ($\Delta L_\text{exp} < 0$). This corresponds to the number of points in the lower left quadrant of \cref{fig:dL_correlation} divided by the number of points in the left half. For the best-performing QIF model, 90\% of steps with an expected loss reduction also reduce the loss, while for the best-performing and asynchronous LIF models the fractions are only 75\% and 62\%, respectively.

\section{Implementation \& Computations}
\label{sec:computation}

Most computations required for this work were implemented using the JAX Python library \cite{jax2018github}, which enables just-in-time compilation and automatic differentiation, and executed on GPU nodes (Nvidia A40, 48GB) on the Marvin HPC cluster of the University of Bonn.
For training models, i.e., the results given in \cref{sec:model_performance}, we usually used a single GPU. Each epoch has a runtime of around 2 to 3 seconds. The total training time for 500 epochs (including data loading and compilation) is typically between 20 and 40 minutes, depending mainly on how many spikes need to be simulated for the specific model.
For landscape generation (\cref{fig:overview,fig:landscapes,fig:jumps}) and probing of intermediate points along the training path (\cref{fig:jumps,fig:fig5,fig:jumps_qif}), we usually used four GPUs in parallel. These tasks may run for multiple hours, depending on the resolution of the grid. By far the most costly computations were the rough and the fine hyperparameter search since they require 17{,}280 and 4{,}800 model trainings, respectively. The total runtime of both searches combined were about 70 days (using a single GPU per training). However, using parallel jobs, this runtime can be reduced significantly. We did not measure detailed memory usage statistics. However, by splitting parallelized tasks into batches, insufficient GPU memory can be easily avoided, at the cost of longer runtimes. For example, when computing the loss and gradient landscapes, we evaluated batches of 20 points in parameter space to avoid GPU memory issues.
Note that performance optimization was not a major focus in this work.

In addition to JAX, we use Optax \cite{deepmind2020jax} for optimization, Chex \cite{deepmind2020jax} for testing utilities, scikit-learn \cite{scikit-learn} for principle component analysis, pandas \cite{mckinney2010data, reback2020pandas} for data management, and Matplotlib \cite{Hunter2007} for visualizations.